\documentclass{article}

\usepackage{arxiv}

\usepackage[utf8]{inputenc} 
\usepackage[T1]{fontenc}    
\usepackage{hyperref}       
\usepackage{url}            
\usepackage{booktabs}       
\usepackage{amsfonts}       
\usepackage{nicefrac}       
\usepackage{microtype}      
\usepackage{lipsum}		
\usepackage{graphicx}
\usepackage[square,sort,comma,numbers]{natbib}
\usepackage{doi}
\usepackage{amssymb}
\usepackage{amsmath}
\usepackage{bm}
\usepackage{amsthm}

\newtheorem{theorem}{Theorem}[section]

\def\E{{\mathbb E}}

\def\0{{\mathbb 0}}
\def\P{{{\mathbb P}}}
\def\Q{{\mathbb Q}}
\def\R{{\mathbb R}}

\def\V{{\mathbb V}}

\def\curlyB{{\mathcal B}}
\def\curlyC{{\mathcal C}}
\def\curlyD{{\mathcal D}}
\def\curlyE{{\mathcal E}}
\def\curlyF{{\mathcal F}}

\def\curlyH{{\mathcal H}}
\def\curlyI{{\mathcal I}}

\def\curlyX{{\mathcal X}}
\def\curlyY{{\mathcal Y}}
\def\curlyZ{{\mathcal Z}}

\makeatletter \renewcommand\d[1]{\ensuremath{%
  \;\mathrm{d}#1\@ifnextchar\d{\!}{}}}
\makeatother

\def\KL{{\mathrm{KL}}}

\title{Variational Autoencoding Neural Operators}

\author{ Jacob H. Seidman \\
Electrical and Systems Engineering\\
Mechanical Engineering and Applied Mechanics \\
University of Pennsylvania \\
\texttt{seidj@sas.upenn.edu}
\AND
Georgios Kissas \\
Mechanical Engineering and Applied Mechanics \\
University of Pennsylvania\\
\texttt{gkissas@seas.upenn.edu}
\AND
George J. Pappas \\
Electrical and Systems Engineering \\
University of Pennsylvania \\
\texttt{pappasg@seas.upenn.edu}
\AND
Paris Perdikaris \\
Mechanical Engineering and Applied Mechanics \\
University of Pennsylvania \\
\texttt{pgp@seas.upenn.edu}
}

\renewcommand{\shorttitle}{Variational Autoencoding Neural Operators}

\begin{document}

\maketitle




\begin{abstract}
Unsupervised learning with functional data is an emerging paradigm of machine learning research with applications to computer vision, climate modeling and physical systems. A natural way of modeling functional data is by learning operators between infinite dimensional spaces, leading to discretization invariant representations that scale independently of the sample grid resolution. Here we present Variational Autoencoding Neural Operators (VANO), a general strategy for making a large class of operator learning architectures act as variational autoencoders. For this purpose, we provide a novel rigorous mathematical formulation of the variational objective in function spaces for training. VANO first maps an input function to a distribution over a latent space using a parametric encoder and then decodes a sample from the latent distribution to reconstruct the input, as in classic variational autoencoders. We test VANO with different model set-ups and architecture choices for a variety of benchmarks. We start from a simple Gaussian random field where we can analytically track what the model learns and progressively transition to more challenging benchmarks including modeling phase separation in Cahn-Hilliard systems and real world satellite data for measuring Earth surface deformation.
\end{abstract}

\section{Introduction}
\label{sec: Intro}

Much of machine learning research focuses on data residing in finite dimensional vector spaces.  For example, images are commonly seen as vectors in a space with dimension equal to the number of pixels \cite{santhanam2017generalized} and words are represented by one-hot encodings in a space representing a dictionary \cite{vaswani2017attention}.  Architectures that act on such data are built with this structure in mind; they aim to learn maps between finite dimensional spaces of data.

On the other hand, physics often models signals of interest in the natural world in terms of continuous fields, e.g. velocity in fluid dynamics or temperature in heat transfer. These fields are typically functions over a continuous domain, and therefore correspond to vectors in \emph{infinite-dimensional} vector spaces, also known as functional data.  To use machine learning tools for continuous signals in these physical applications, models must be able to act on and return representations of functional data.

The most straightforward way to do this is known as the discretize-first approach.  Here, functional data is mapped into a finite dimensional vector space via measurements along a predefined collection of locations.  At this point, standard machine learning tools for finite dimensional data can be used to generate measurements of a desired output function, also evaluated along a predefined set of locations.  The drawback of these methods is their rigidity with respect to the underlying discretization scheme; they will not be able to evaluate the output function at any location outside of the original discretization.

As an alternative, operator learning methods aim to design models which give well defined operators between the function spaces themselves instead of their discretizations.  These methods often take a discretization agnostic approach and are able to produce outputs that can be queried at arbitrary points in their target domain.  The Graph Neural Operator \cite{anandkumar2020neural} proposed a compositional architecture built from parameterized integral transformations of the input combined with point-wise linear and nonlinear maps.  This approach was modified to leverage fast Fourier transforms in computing the integral transform component, leading to the Fourier Neural Operator \cite{li2020fourier}, U-net variants \cite{wen2022u}, as well as guarantees of universal approximation  \cite{kovachki2021universal}.  

Inspired from one of the first operator architectures in \cite{chen1995universal}, the DeepONet \cite{lu2021learning} uses finite dimensional representations of input functions to derive coefficients along a learned basis of output functions.  While this approach has been shown to have universal approximation properties as well \cite{lanthaler2022error}, the required size of the architecture can scale unfavorably due to the linear nature of the output function representations \cite{lanthaler2022error, seidman2022nomad}.  Under the assumption that the output functions concentrate along a finite dimensional manifold in the ambient function space, \cite{seidman2022nomad} proposed a model which builds nonlinear parameterizations of output functions and circumvents the limitations of purely linear representations.

While much recent work has focused on designing methods for functional data in a supervised setting, less has been done for  unsupervised learning.  Here we focus on two key aspects of unsupervised learning, namely dimensionality reduction and generative modeling.  For dimensionality reduction of functional data living in a Hilbert space, the inner product structure allows for generalizations of principal components analysis (PCA) \cite{wang2016functional}, also known as proper orthogonal decomposition \cite{chatterjee2000introduction}.  Kernel tricks can also be employed on functional data to obtain nonlinear versions of PCA in feature spaces \cite{song2021nonlinear}.  A generalization of the manifold learning \cite{nadler2006diffusion} approach was taken in \cite{du2021learning} to learn distance preserving and locally linear embeddings of functional data into finite dimensional spaces. 

Generative modeling of functional data has been approached by defining stochastic processes with neural networks, dubbed neural processes \cite{garnelo2018neural, garnelo2018conditional, kim2018attentive}.  Adversarial generative models for continuous images trained on point-wise data have also been proposed in \cite{skorokhodov2021adversarial} and \cite{dupont2021generative}, while a variational autoencoder (VAE) approach with neural radiance fields (NeRFs) was taken in \cite{kosiorek2021nerf}.  

These methods formulate their training objective in terms of point-wise measurements, resulting in models which learn to maximize the probability of observing a collection of points and not a function itself. This makes the function sampling density play a decisive role in how well the model performs; if high and low resolution data coexist in a data-set, the model will over-fit the high resolution data \cite{rahman2022generative}. Recently, the U-net FNO architecture \cite{wen2022u} was used in the Generative Adversarial Neural Operator (GANO) framework \cite{rahman2022generative} to build a generator acting on samples of Gaussian random fields and a functional discriminator in order to overcome the aforementioned drawbacks.

\begin{figure*}[t]
\begin{center}
\includegraphics[width=0.75\textwidth]{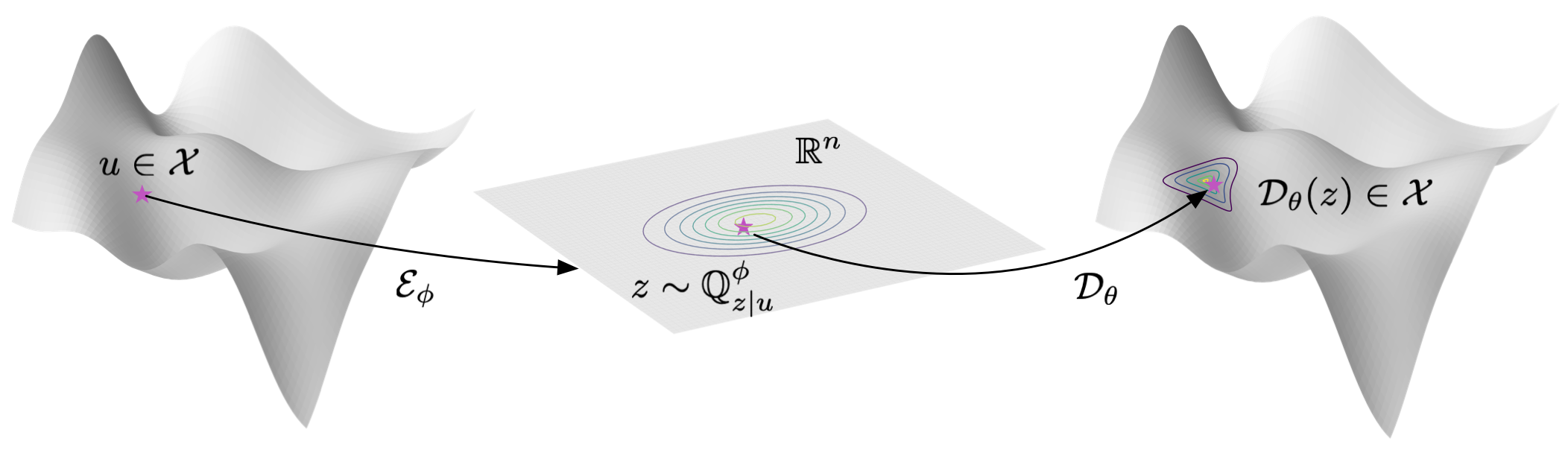}
\caption{Schematic representation of the VANO framework: The encoder $\mathcal{E}_\phi$ maps a point from the input function manifold to a random point sampled from a variational distribution $\mathbb{Q}^\phi_{z|u}$ which is then mapped to a point on the output function manifold using the decoder $\mathcal{D}_\theta$.}
\label{fig:master}
\end{center}
\end{figure*}

In this paper, we provide a novel method of using encoder-decoder operator learning architectures for dimensionality reduction and generative modeling of functional data.  Under the manifold hypothesis for functional data \cite{seidman2022nomad}, if we train these models to learn the identity map then the latent space attempts to learn finite dimensional coordinates for the data manifold.  In Figure \ref{fig:master} we give a visual representation of our approach.  We will additionally view the data as coming from a generative model on this coordinate space and train the corresponding operator learning architecture via an auto-encoding variational Bayes approach.  The decoder component of the architecture then creates functions from samples in the latent coordinate space which can be queried at any point along their domain.  To overcome over-fitting pathologies at higher data resolutions we describe the variational objective in a \emph{discretization agnostic way} by putting forth a well-defined formulation with respect to the functional data space instead of spaces of point-wise evaluations.

Our main contributions can be summarized as:
\begin{itemize}
\item We provide the first rigorous mathematical formulation of a variational objective which is completely discretization agnostic and easily computable.
\item Using this objective, we give a novel formulation of a variational autoencoder for functional data with operator learning architectures.
\item We perform zero-shot super-resolution sampling of functions describing complex physical processes and real world satellite data.
\item We demonstrate state-of-the-art performance in terms of reconstruction error and sample generation quality while taking a fraction of the training time and model size compared to competing approaches.
\end{itemize}
\paragraph{Outline of paper:} The remainder of the paper will be structured as follows.  First we will describe the class of encoder-decoder operator learning architectures and describe how they perform dimensionality reduction on functional data.  Next, we will briefly review the VAE formulation for finite dimensional data before giving a mathematically precise generalization to functional data and describing the components of our model.  Section \ref{sec: experiments} will present several experiments illustrating the effectiveness of this approach. 


\paragraph{Notation}
We denote our functional data by $u \in \curlyX$, where $\curlyX$ is a function space over some domain $X \subset \R^{d}$.  Typically we will take $\curlyX = L^2(X)$ or $\curlyX = H^s(X)$, where $H^s(X)$ is the Sobolev space of index $s$ over $X$.  Probability measures will be denoted by blackboard bold typefaced letters $\P$, $\Q$, and $\V$.

\section{Encoder-Decoder Neural Operators}
A large class of architectures built to learn mappings between spaces of functions $\curlyX \to \curlyY$ make use of a finite dimensional latent space in the following way.  First, an encoding map $\curlyE:\curlyX \to \R^n$ is learned from the input functions to a $n$-dimensional latent space. Then, the latent code corresponding to a functional input, $z = \curlyE(u)$ is mapped to a queryable output function via a decoding map $\curlyD: \R^n \to \curlyY$, $f(\cdot) = \curlyD(z)$. For example, in the DeepONet architecture \cite{lu2021learning}, input functions $u$ are mapped via a ``branch network'' to a hidden vector $z \in \R^n$, which is then used as coefficients of a learned basis to reconstruct an output function.  These methods can be interpreted as giving a finite dimensional parameterization to the set of output functions, where the parameters for a fixed output function are determined from the corresponding input function.  

\subsection{Linear versus Nonlinear Decoders}
It was shown in \cite{seidman2022nomad} that when these kinds of architectures build output functions in a linear manner from the latent space, such as in \cite{kissas2022learning, lu2021learning, bhattacharya2021model}, they may miss low dimensional nonlinear structure in the set of output functions that can otherwise be captured by a nonlinear map from the latent space to the output function space.  The authors further gave an interpretation of this architecture under the assumption that the distribution of output functions concentrates on a low dimensional manifold in its ambient function space.  In this setting, the decoder map ideally would learn a coordinate chart between the finite dimensional latent space and the manifold of output functions.  This suggests that successful architectures are implicitly performing dimensionality reduction on the set of output functions.

\subsection{Dimensionality Reduction through the Latent Space}

We will follow this interpretation to create a natural extension of encoder-decoder operator learning architectures for dimensionality reduction and generation of functional data.  If the input and output function spaces are the same $\curlyX = \curlyY$ and we learn the identity map on our data factored through a finite dimensional latent space, then the encoding map $\curlyE: \curlyX \to \R^n$ gives a lower dimensional representation of our functional data.  That is, when trained to approximate the identity map, these architectures become {\em functional autoencoders}.

If we additionally would like to generate new samples of our functional data with this framework, it would suffice to learn a probability measure over the latent space corresponding to the finite dimensional embedding of our data.  Similar to the non-functional data case, this can be modelled through the use of a Variational Auto-Encoder (VAE) \cite{kingma2013auto}, which takes a Bayesian approach to determining latent representations of observed  data.  While this method has been studied extensively on finite dimensional data, its extension to functional data has only been explored specifically for neural radiance fields in \cite{kosiorek2021nerf}, where the variational objective is formulated in terms of point-wise measurements.

In this paper, we will place operator learning architectures with a finite dimensional latent space, such as DeepONet \cite{lu2021learning} and NOMAD \cite{seidman2022nomad}, within the formalism of autoencoding variational Bayesian methods to simultaneously obtain a new method of dimensionality reduction and generative modeling for functional data.  To do so, we must be careful to reformulate the VAE objective in function spaces.  Variational objectives have been formulated considering latent spaces as function spaces, as in \cite{wild2021variational, wild2022generalized}, but a variational objective where the likelihood term is described in a functional data space has not yet been addressed.

As we will see, while the immediate application of the formulation for finite dimensional data does not apply, there exists an appropriate generalization which is mathematically rigorous and practically well behaved.  The benefit of the function space formulation is the lack of reference to a particular choice of discretization of the data, leading to a more flexible objective which remains valid under different measurements of the functional data.

\section{VAEs for Finite Dimensional Data}

Here we review a simple generative model for finite dimensional data and the resulting variational Bayesian approach to inference in the latent space.  For this subsection only, we will define our data space as $\curlyX = \R^d$.  As before, let the latent space be $\curlyZ = \R^n$, often with $n << d$. 

Assume the following generative model for samples of $u$ from its probability measure $\P_u$ on $\curlyX$.  Let $\P_z$ be a \emph{prior} probability measure on $\curlyZ$, $\curlyD: \curlyZ \to \curlyX$ a function from the latent space to the data space, and $\eta$ a noise vector sampled from a probability measure $\V$ on $\curlyX$ such that
\begin{equation} \label{eq: gen model}
u = \curlyD(z) + \eta, \quad z \sim \P_z, \quad \eta \sim \V,
\end{equation}
is distributed according to $\P_u$.

According to this model, there exists a joint probability measure $\P$ on $\curlyZ \times \curlyX$ with marginals $\P_z$ and $\P_u$ as defined above.  Assume these three measures have well defined probability density functions, $p(z,u)$, $p(z)$ and $p(u)$, respectively, and conditional densities $p(z|u)$ and $p(u|z)$.

Under full knowledge of these densities, we can form a low dimensional representation of a given data point $u$ by sampling from $p(z|u)$.  However, in general, the evaluation of the conditional density is intractable.  This motivates the variational approach \cite{kingma2013auto}, where we instead create a parameterized family of distributions $q^\phi(z|u)$, and attempt to approximate the true conditional $p(z|u)$ with $q^\phi(z|u)$.  When the KL divergence is used as a quality of approximation from $q^\phi(z|u)$ to $p(z|u)$, this can be approached with the following optimization problem
\begin{equation} \label{eq: KL variational conditional ob}
\underset{\phi}{\mathrm{minimize}}\quad \underset{u \sim p(u)}{\E} \left[ \KL[q^\phi(z|u) \;||\; p(z|u)]\right].
\end{equation}
Since we do not have access to $p(z|u)$ and cannot evaluate the KL divergence term above, we optimize instead a quantity known as the \emph{Evidence Lower Bound} (ELBO),
\begin{equation} \label{eq: fd elbo}
\begin{split}
\mathcal L = -\underset{z \sim q^{\phi}(z|u)}{\E}[\log p(u|z)] + \KL[q^\phi(z|u)\;||\;p(z)],
\end{split}
\end{equation}
which differs from the objective in \eqref{eq: KL variational conditional ob} by a data-dependent constant,
\begin{equation}
\KL[q^\phi(z|u) \;||\; p(z|u)] = -\mathcal L + \log p(x)
\end{equation}
When the prior $p(z)$ is a Gaussian and the variational distribution $q^\phi(z|u)$ is also Gaussian with a mean and variance dependent on the input $u$ through a parameterized encoding map $\curlyE_\phi(u) = (\mu_\phi(u), \sigma_\phi(u)))$, the KL term in \eqref{eq: fd elbo} has a closed form.

Under the assumption that the noise vector in \eqref{eq: gen model} is a centered Gaussian with isotropic covariance, $\eta \sim \mathcal N(0, \delta^2 \mathrm{I})$, the log likelihood term is equal to
\begin{equation} \label{eq: fd likelihood}
\log p(u\;|\;z) = \log (2\pi\delta^2)^{-d/2} -\frac{1}{2\delta^2}\|u - \curlyD(z)\|_2^2.
\end{equation}
By parameterizing the function $\curlyD$ as well, we arrive at an objective function that can be used to train the encoder $\curlyE_{\phi}$ and decoder $\curlyD_{\theta}$ in an end-to-end fashion.

\section{VAEs for Functional Data}

We begin formulating a VAE in this case analogously to the previous section; we posit the generative model in \eqref{eq: gen model} which induces a joint measure $\P$ on $\curlyZ \times \curlyX$ with marginals $\P_u$ and $\P_z$.  Under mild assumptions on the spaces $\curlyZ$ and $\curlyX$ (such as being a separable Banach spaces), there exist \emph{regular conditional measures} $\P_{z|u}$ and $\P_{u|z}$ which are well defined $\P_u$-a.e. and $\P_z$-a.e., respectively.

At this point the formulation begins to diverge from the finite dimensional case.  In an infinite dimensional function space, such as $\curlyX$, we no longer have a canonical notion of a probability density function to formulate our objective function and ELBO.  In particular, the first term of \eqref{eq: fd elbo} is no longer well defined as written.  We will instead reason in terms of the probability measures $\P_u$, $\P_z$, $\P_{z|u}$, $\P_{u|z}$, the variational family of measures $\Q^\phi_{z|u}$, the noise process measure $\V$, and various Radon-Nikodym derivatives between them.  Proceeding in this manner we are able to derive the appropriate generalization of \eqref{eq: fd elbo} for data in the function space $\curlyX$.

\begin{theorem} \label{thm: fELBO}

Let $\curlyX$ and $\curlyZ$ be Polish spaces. Given the generative model \eqref{eq: gen model}, assume that the conditional measure $\P_{u|z}$ is absolutely continuous with respect to the noise measure $\V$. Then the following holds
\begin{equation} 
 \KL[\Q^\phi_{z|u}\;||\;\P_z] = -\mathcal L + \log \frac{\d \P_u}{\d \V}.
\end{equation}
with
\begin{equation} \label{eq: fELBO}
\mathcal L = - \underset{z \sim \Q^{\phi}_{z|u}}{\E}\Big[\log \frac{d \P_{u|z}}{d\V}(u)\Big] + \KL[\Q^\phi_{z|u}\;||\;\P_z]
\end{equation}
\end{theorem}
\begin{proof}
The proof is provided in Appendix \ref{thm: fELBO}.
\end{proof}

The benefit of this formulation is that the objective function makes no reference to any particular choice of discretization or available function measurements.  In this sense, it is a training objective that is truly defined on a function space; whatever measurements are available will be used to approximate this objective, ensuring a form of consistency over varying discretization schemes.

\subsection{Computing the ELBO Objective}

To compute the likelihood term in \eqref{eq: fELBO}, first note that under the generative model \eqref{eq: gen model}, given $z \in \curlyZ$ the conditional measure $\P_{u|z}$ corresponds to a shifted version of the noise process centered at $\curlyD(z)$.  The Radon-Nikodym derivative $\frac{\d\P_{u|z}}{\d\V}$ then represents the change of measure of $\V$ under a shift by $\curlyD(z)$.

In this work we will assume that $\V$ is a \emph{Gaussian} measure on the space $\curlyX$.  Changes of measure for translated Gaussian measures are well understood and are described by the \emph{Cameron-Martin} formula (see Appendix \ref{ap: gaussian measures}); this will be the main tool which allows us to evaluate the first term in \eqref{eq: fELBO}.

In particular, we will take $\eta$ to be a pure white noise process and $\V$ the corresponding white noise measure. Note that this implies that our measured signals must live in the dual Sobolev space $\curlyX = H^{-s}(X)$ for any $s > d/2$ \citep{lasanen2018elliptic}.  In this case, the Cameron-Martin formula gives
\begin{equation} \label{eq: CM white noise likelihood}
\log \frac{\d\P_{u|z}}{\d\V}(u) = -\frac{1}{2}\|\curlyD(z)\|_{L^2}^2 - \langle \curlyD(z), u \rangle^{\sim},
\end{equation}
where $\langle D(z), u \rangle^{\sim}$ can be thought of as the inner product on $L^2(X)$ extended to $H^{-s}(X)$ in the second argument and is well defined a.e. with respect to the noise process $\V$.  Given sensor measurements of $u$, we can approximate this second term as the standard inner product with the corresponding measurements of $\curlyD(z)$.  For more details on Gaussian measures in Banach spaces, white noise, and the Cameron-Martin formula see Appendix \ref{ap: gaussian measures}.

Note that the expression in \eqref{eq: CM white noise likelihood} is the same as
\begin{equation*}
-\frac{1}{2}\|\curlyD(z) - u\|_{L^2}^2 = -\frac{1}{2}\|\curlyD(z)\|_{L^2}^2 - \langle \curlyD(z), u \rangle - \frac{1}{2}\|u\|_{L^2}^2,
\end{equation*}
except for the last term.  This is what we would expect to see when our data is not functional and lies in $\R^d$ with a Gaussian likelihood, but since $u$ is drawn from a shifted white noise measure it is not in $L^2$.  However, we see that the expression we derived instead for the likelihood is the same as what we would like to use up to the model independent term $\|u\|_{L^2}^2$.  In this sense, the white noise likelihood formulation of the ELBO is the natural extension of the Gaussian likelihood from the finite dimensional case.

\begin{figure*}
\begin{center}
\centerline{\includegraphics[width=.9\textwidth]{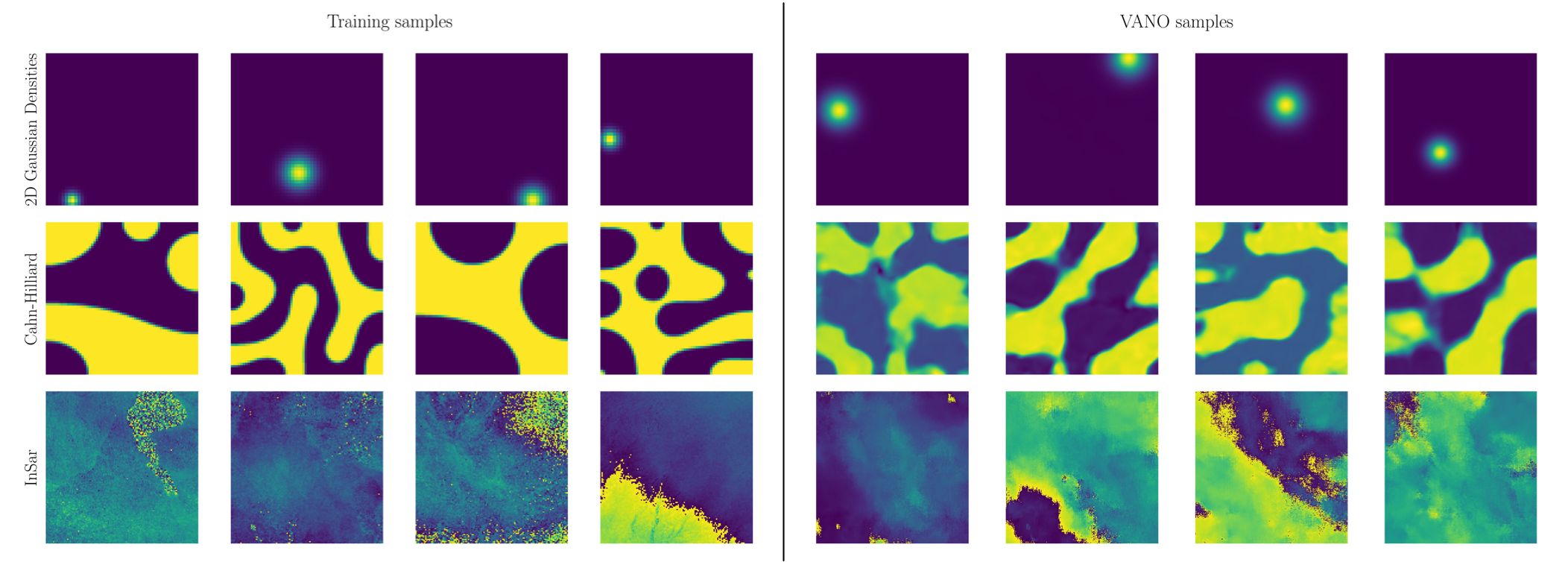}}
\caption{Representative function samples from the different benchmarks considered in this work. Left: Example functions from the testing data-sets, Right: Super-resolution samples generated by VANO.}
\label{fig:experiment_samples}
\end{center}
\end{figure*}

\section{Variational Autoencoding Neural Operators}


Given Theorem \ref{thm: fELBO} we can define the full Variational Autoencoding Neural Operator (VANO) after making choices for the encoding and decoding maps. See Figure \ref{fig:master} for a visual illustration of the overall architecture.
\paragraph{Encoder:} The encoder will map a function $u \in \curlyX$ to the probability measure $\Q^\phi_{z|u}$ on the latent space $\R^n$.  We choose the variational family $\Q^{\phi}_{z|u}$ to be multivariate Gaussians with diagonal covariance.  It then suffices that the encoding map takes as input the function $u$, and returns a mean $\mu(u) \in \R^n$ and $n$ positive scalars $\sigma_1, \ldots, \sigma_n = \sigma \in \R^n$ to parameterize this Gaussian.  Hence, we define the encoder as a map $\curlyE^\phi: \curlyX \to \R^n \times \R^n$.  In this paper, we will use architectures which pass measurements of the input function $u$ through a neural network of fixed architecture.  These measurements can either be point-wise, as we take to be the case in this paper, but could also be projections onto sets of functions such as trigonometric polynomials, wavelets, or other parameterized functions.  
\paragraph{Decoder:} The decoder will take a sample $z$ of a probability measure on the latent space $\R^n$ and map it to a function $\curlyD(z) \in \curlyX$ that can be queried at any point.  In this paper, we will parameterize decoders by defining a neural network which takes in points in the domain of the functions in $\curlyX$, and condition its forward pass on the latent variable $z$.  Here we will use two main variations of this conditioning process: linear conditioning, and concatenation conditioning (see Appendix \ref{sec:dec choices} for details).
\paragraph{Evaluating ELBO for Training:} Given a data-set of $N$ functions $\{u^i\}_{i=1}^N$, we train VANO by optimizing the objective function
\begin{equation} \label{eq: training obj}
\mathcal L(\phi, \theta) = \frac{1}{N} \sum_{i=1}^N \Big[\E_{{\Q^\phi_{z|u^i}}}\left[\frac{1}{2} \|\curlyD_\theta(z)\|_{L^2}^2 - \langle \curlyD_\theta(z), \!u ^i\rangle^{\sim}\right] + \KL[\Q^\phi_{z|u^i}\;||\;\P_z]\Big].
\end{equation}
The expectation over the posterior $\Q^\phi_{z|u^i}$ is approximated via Monte-Carlo by sampling $S$ latent variables $z \sim \Q^\phi_{z|u^i}$ and computing an empirical expectation.  The re-parameterization trick \cite{kingma2013auto} is used when sampling from $\Q^{\phi}_{z|u^i}$ to decouple the randomness from the parameters of the encoder $\curlyE^\phi$ and allow for the computation of gradients with respect to the parameters $\phi$.  As $\curlyD(z)$ can be evaluated at any point in the domain $X$, we can approximate the terms inside this expectation with whichever measurements are available for the data $u^i$.  For example, with point-wise measurements $u(x_1), \ldots, u(x_m)$ we can use the approximation
\begin{equation}
\langle \curlyD_\theta(z), u ^i\rangle^{\sim} \approx \sum_{i=1}^m \curlyD_\theta(x_i)u(x_i).
\end{equation}

To avoid pathologies in the optimization of \eqref{eq: training obj}, we train our models with a scalar hyper-parameter $\beta$ in front of the KL divergence term as in \cite{higgins2017betavae} to balance the interplay between the KL and reconstruction losses.   

\section{Experiments} \label{sec: experiments}
In this section we consider four examples for testing the performance of our model.  In the first example, we learn the distribution corresponding to a Gaussian random field (GRF).  We show that a model with a linear decoder architecture is able to accurately recover the Karhunen-Lo\'eve decomposition \cite{adler1990introduction}, which is known to be an $L^2$ optimal dimension reduced approximation of the true field.  Next, we examine the impact of different decoder architectures for learning distributions which do not immediately concentrate on low dimensional linear spaces, using a functional data-set of bivariate Gaussian pdfs. Next, we learn solutions to a Cahn-Hilliard partial differential equation (PDE) system representing patterns from a phase separation processes in binary mixtures.  Finally, we employ the real world InSAR interferogram data-set presented in \cite{rahman2022generative} to demonstrate state-of-the-art performance compared to recent operator learning methods for generative modeling. For the first three experiments, we use a MMD metric for functional data \cite{wynne2022kernel} between samples from the different models and the ground truth to assess performance. For the last example we compare angular statistics between model generated samples and the ground truth. As an overview, Figure \ref{fig:experiment_samples} shows some samples generated by our model on three of the data-sets we consider.  More details on the metrics, hyper-parameters, architectures and the training set-up can be found in the Appendix in Section \ref{sec:exp details}.

\subsection{Gaussian Random Field}
The motivation of this example is to study the quality of the reduced dimension approximation that our model learns. For this purpose, we aim to learn a zero mean Gaussian random field (GRF) on $X = [0,1]$ with zero boundary conditions and covariance operator $\Gamma = (\mathrm{I} - \Delta)^{-\alpha}$.  This operator admits the orthonormal eigendecomposition
\begin{align*}
\Gamma= \sum_{i=1}^\infty \lambda_i \varphi_i \otimes \varphi_i,
\end{align*}
where $\lambda_i = ((2 \pi i)^2  + \tau^2)^{-\alpha}$ and $\varphi_i(x)= \sqrt{2} \sin(2 \pi i x)$.  From the Karhunen-Lo\'eve theorem \citep{adler1990introduction} we can construct random functions distributed as
\begin{align} \label{eq: KL mu}
u = \sum_{i=1}^\infty \xi_i \sqrt{\lambda_i} \varphi_i,
\end{align}
where $\xi_i \sim \mathcal{N}(0,1)$ are normally distributed random variables.

We use the above sum truncated at 32 eigenpairs to construct a data-set of $N$ functions $\{u^i\}_{i=1}^N$ and use it to train a Variational Autoencoding Neural Operator with a linear decoder.  By setting the prior $\P_z$ to be a standard Gaussian on $\R^n$, a linear decoder $\curlyD$ learns basis functions $\tau_i \in \curlyX$ which map samples from the prior to functions,
\begin{equation*} \label{eq: linear gen}
\curlyD(z)(x) = \sum_{i=1}^{n} z_i \tau_i(x), \quad z_i\sim \mathcal{N}(0,1), \quad \text{i.i.d.}
\end{equation*}
The Karhunen Lo\'eve theorem again tells us that the optimal choice of decoder basis functions $\tau_i$ should be exactly the eigenfunctions $\varphi_i$ of the covariance operator $\Gamma$ scaled by $\sqrt{\lambda_i}$. To evaluate the model performance we compare the covariance operators between the true Gaussian random field and the learned model \eqref{eq: linear gen} using a normalized Hilbert-Schmidt norm,
$\|\Gamma - \hat \Gamma\|_{HS}^2/\|\Gamma\|_{HS}^2$.

\begin{figure}
\begin{center}
\centerline{\includegraphics[width=\columnwidth]{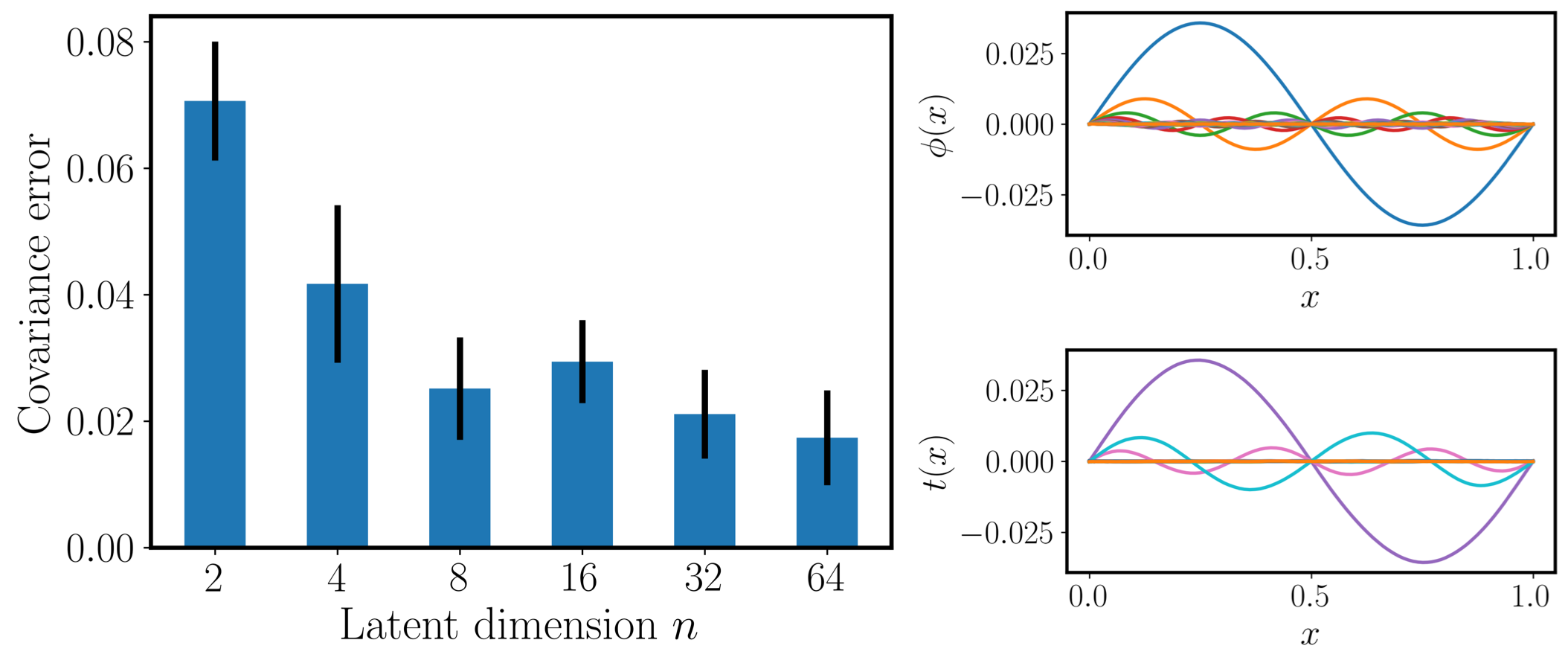}}
\caption{Left: Normalized Hilbert-Schmidt norm error between GRF samples generated from VANO and the ground truth for different sizes of the latent space.  Right: Comparison between the optimal basis from the Karhunen Lo\'eve theorem (top) and the learned basis (bottom).}
\label{fig:grf1d_sweep_boxplot}
\end{center}
\end{figure}

We present the values of the normalized Hilbert-Schmidt norm for different latent dimension sizes and over multiple model initializations in Figure \ref{fig:grf1d_sweep_boxplot}.  The right side of Figure \ref{fig:grf1d_sweep_boxplot} shows that the learned basis functions align closely with the optimal choice from the Karhunen Lo\'eve theorem, scaled eigenfunctions $\sqrt{\lambda_i}\phi_i$.  Generated samples from the trained model are depicted in Figure \ref{fig:generated_samples_comparison} in the Appendix. 

\subsection{The Need for Nonlinear Decoders}

In this example we examine the effect of using linear versus nonlinear decoders for learning distributions of functional data.  We construct a data-set consisting of bivariate Gaussian density functions over the unit square $[0,1]^2$ where the mean is sampled randomly within the domain and the covariance is a random positive multiple of the identity.  Performing PCA on this data-set shows a spectrum of eigenvalues with slow decay, indicating that architectures with linear decoders will not be able to capture these functional data unless their hidden dimension is very large (see Appendix figure \ref{fig:GaussianBump/evalues_decay_gaussian}). To test this, we train VANO models with a linear and nonlinear decoder, respectively, over a range of hidden dimensions.  We measure the distance of the learned distribution of functions to the ground truth via the generalized MMD distance, see Appendix \ref{sec: generalized mmd}.

\begin{figure}
\begin{center}
\centerline{\includegraphics[width=\columnwidth]{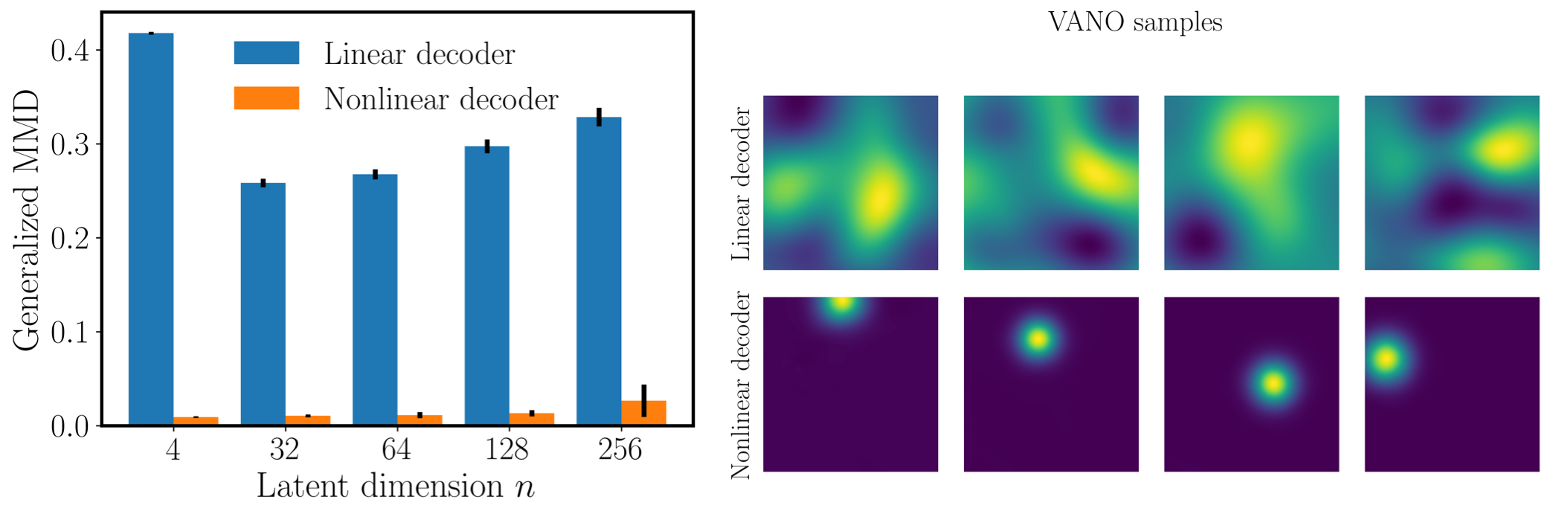}}
\caption{Left: Generalized MMD values for VANO with a linear and a nonlinear decoder for different latent dimensions, Right: Generated samples from VANO with a linear (top) and with a nonlinear decoder (bottom).}
\label{fig:GaussianBump/bumps2d_sweep}
\end{center}
\end{figure}

In the left panel of Figure \ref{fig:GaussianBump/bumps2d_sweep} we present values of the MMD metric over five individual runs and samples generated from VANO with a linear and nonlinear decoder. We observe that the error metric for the linear model takes very high values even at larger latent dimensions. In the right panel we see representative samples generated from linear and nonlinear decoder versions of VANO.  The linear decoder is not able to localize the 2D functions as in the data-set, while the nonlinear decoder produces very similar samples. In Figures \ref{fig:GaussianBump/test_reconstruction_linear} and \ref{fig:GaussianBump/test_reconstructions} of the Appendix we show the reconstruction of randomly chosen cases from the test data-set for the linear and nonlinear decoders, respectively. More samples from the linear and nonlinear decoder versions of VANO are shown in Figures \ref{fig:GaussianBump/samples_gaussian_linear} and \ref{fig:GaussianBump/gaussian_density_samples} of the Appendix.

\subsection{Phase separation patterns in Cahn-Hilliard systems}

As a more challenging benchmark we consider the Cahn-Hilliard patterns data-set \cite{kobeissi2022mechanical} which contains different patterns derived from the solution of the Cahn-Hilliard equation. The Cahn-Hilliard equation is a fourth-order partial differential equation that describes the evolution of the phase separation process in binary material mixtures, see Appendix Section \ref{sec:Cahn-Hilliard Appendix}, for details.

For this problem we compare the VANO model with a discretize-first convolutional VAE approach over different resolutions. We train the discrete VAE on images of different resolutions namely 64x64, 128x128 and 256x256 and compare against the proposed VANO model which we train only on 64x64 images and generate samples at higher resolutions. We present the results of the generalized MMD metric for each model in Table \ref{tab:vano_vae}. 

We observe that the VANO model performs better than the VAE at resolution 64x64, and slightly worse at generating samples at resolutions 128x128 and 256x256.  However, even though the VANO model was not trained on these resolutions, it still gives competitive performance to the discretize-first method which did have access to high resolution data during training.  Additionally, the size of the discretize-first model increases with the desired resolution while the size of the VANO remains fixed.  This highlights the super-resolution capabilities of using models designed to output true functional data.  Additionally, we present functions sampled from both models in the Appendix Section \ref{sec:InSAR Appendix}, as well as reconstructions from each model at different resolutions.  In Figure \ref{fig:CahnHilliard/samples_comparisons_2} we see that at higher resolutions the samples generated by the VANO model have smoother boundaries and appear more natural than those created by the discretize-first VAE approach. 

\begin{figure*}
\begin{center}
\centerline{\includegraphics[width=\textwidth]{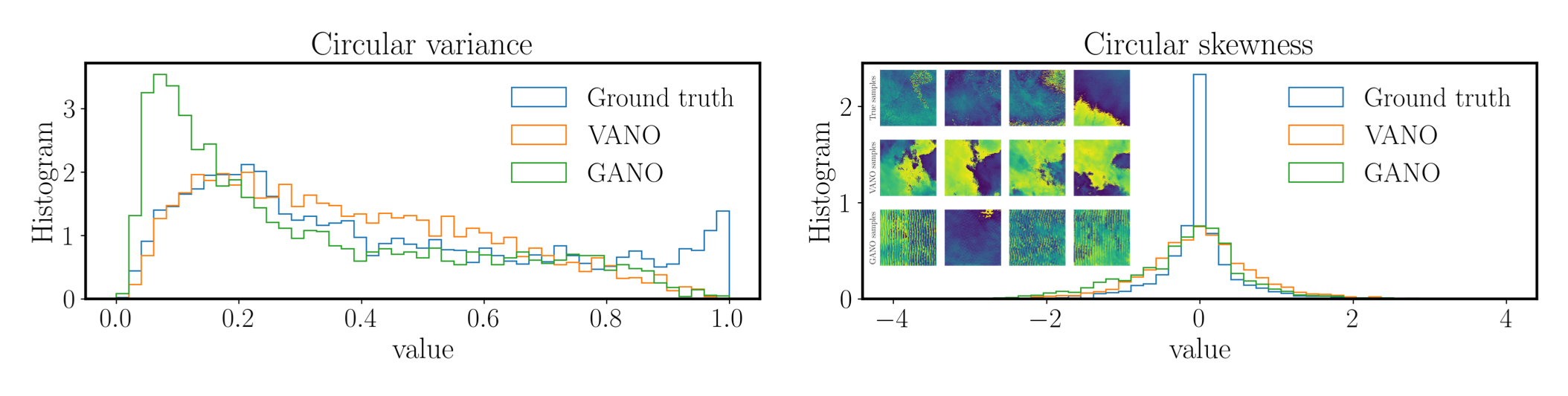}}
\caption{Left: Circular variance of the true data-set, the VANO generated data and the GANO generated data. Right: Circular skewness and generated samples between the true data-set, the VANO generated data and the GANO generated data.}
\label{fig:Volcano/volcano_statistics}
\end{center}
\end{figure*}

\begin{table}
\caption{Generalized MMD distance between ground truth test samples and samples generated by different models. Subscripts indicate the resolution of the data used to train each model.}
\label{tab:vano_vae}
\begin{center}
\begin{small}
\begin{sc}
\begin{tabular}{lcccr}
\toprule
MMD$\times10^{-2}$  & $64\times 64$ & $128\times 128$ & $256\times 256$ \\
\midrule
VANO$_{64\times 64}$  & 0.71 $\pm$ 0.04 & 0.88 $\pm$ 0.04 & 1.34 $\pm$ 0.06 \\
VAE$_{64\times 64}$   & 0.73 $\pm$ 0.02 & - & - \\
VAE$_{128\times 128}$ & -  & 0.64 $\pm$ 0.03 & - \\
VAE$_{256\times 256}$ & - &  - &  1.03$\pm$ 0.03 \\
\bottomrule
\end{tabular}
\end{sc}
\end{small}
\end{center}
\end{table}

\subsection{Interferometric Synthetic Aperture Radar data-set}

As a final example, we consider the data-set proposed by Rahman et. al. \cite{rahman2022generative} consisting of Interferometric Synthetic Aperture Radar (InSAR) data. InSAR is a sensing technology that exploits radar signals from aerial vehicles to measure the deformation of the Earth surface for studying the dilation of volcanoes, earthquakes or underwater reserves. As a comparison, we use the Generative Adversarial Neural Operator (GANO) architecture and training parameters provided in \cite{rahman2022generative}.

We train VANO on the entire data-set using the set-up provided in the Appendix Section \ref{sec:InSAR Appendix}. We evaluate the performance of our model using two metrics: circular variance and circular skewness. These are moments of angular random variables, see \cite{rahman2022generative}, used to evaluate the quality of the generated functions. In Figure \ref{fig:Volcano/volcano_statistics} we present a comparison between the circular statistics \cite{rahman2022generative}, see Appendix Section \ref{sec: circular metrics} for details on the metrics, for $N=4096$ samples from the true data-set, and those created from VANO and GANO.  In Section \ref{sec:InSAR Appendix} we present samples generated from both models. We observe that the VANO model achieves superior performance both in terms of circular statistics metrics, as well as in generating realistic samples without spurious artifacts.  In Figure \ref{fig:Volcano/test_reconstructions} we present sample reconstructions of the data from VANO and see that it also acts as a denoiser for the original data.  Moreover, we find that VANO is trains 4x faster, with the 1/4 of model size compared to GANO (see Appendix Tables  \ref{tab:computational_cost} and \ref{tab:number_of_parameters}).

\section{Discussion}

In this work, we have shown that a large class of architectures designed for supervised operator learning can be modified to behave as variational auto-encoders for functional data.  The performance of this approach was demonstrated through learning generative models for functional data coming from synthetic benchmarks, solutions of PDE systems, and real satellite data.  By deriving an appropriate variational objective in an (infinite-dimensional) functional data space, we placed this approach on firm mathematical footing.

These models inherit some limitations common to all VAE approaches.  In particular, there is a constant tension in the objective function of balancing the reconstruction loss of the data and the distance of the variational posterior distribution to the prior.  The additional scalar parameter $\beta$ multiplying the KL divergence term \cite{higgins2017betavae} attempts to provide some control of this balance, but the performance of the model can be sensitive to the setting of this parameter.  Some insight can be gained into controlling this phenomenon through rate distortion theory \cite{burgess2018beta}, and it is an interesting direction of future work to generalize this to functional data.  The choice of prior distribution on the latent space can also have a large impact on model performance.  Hierarchical priors could provide additional structure for the generative model, as has been shown in \cite{vahdat2020nvae}.

The approach presented in this paper bears some similarity to generative models built for learning neural fields in computer vision \cite{chen2019learning, anokhin2021image}.  Our theoretical foundation of a variational lower bound for functional data can be directly applied to these approaches instead of the typical formulation of a likelihood on fixed point-wise measurements.  This similarity also points to a larger connection between operator learning methods and conditioned neural fields in vision applications \cite{xie2022neural}.  Both approaches aim to build neural representations of functions which can be queried at arbitrary points of their domains and the techniques developed to do so are likely to be useful across both domains.

Finally, adapting the conditional version of the variational objective \cite{sohn2015learning} can be useful for supervised operator learning where there is aleatoric uncertainty in the output functions.  For example, this can often be the case in inverse problems where not all functional outputs of a system are available for observation.  Applying the variational approach put forth in this paper would give a natural notion of uncertainty quantification, while retaining the approximation capabilities of encoder-decoder operator learning architectures.

\section*{Acknowledgments}
We would like to acknowledge support from the US Department of Energy under under the Advanced Scientific Computing Research program (grant DE-SC0019116) and the US Air Force Office of Scientific Research (grant AFOSR FA9550-20-1-0060). We also thank the developers of the software that enabled our research, including JAX \citep{jax2018github}, Matplotlib \citep{hunter2007matplotlib}, Pytorch \citep{paszke2019pytorch} and NumPy \citep{harris2020array}.

\bibliography{bibliography}
\bibliographystyle{unsrt}

\newpage
\appendix
\onecolumn



\section{Review of Gaussian Measures} \label{ap: gaussian measures}

Here we review some basic facts on Gaussian Measures defined Banach spaces. Our presentation follows the discussion in \cite{kuo1975gaussian, bogachev2015gaussian} and Chapter 8 of \cite{stroock2010probability}.  We call a probability measure $\P$ on the Borel $\sigma$-algebra $\curlyB(\curlyX)$ a \emph{Gaussian measure} if for every $f \in \curlyX^*$, the pushforward measure $\P \circ f^{-1}$ is Gaussian on $\R$.  For such a measure there exists two objects that completely characterize it.  The \emph{mean} is an element $m \in \curlyX$ such that for all $f \in \curlyX^*$,
\begin{equation}
(f, m) = \int_\curlyX (f,x)\d\P(x),
\end{equation}
and the covariance operator $\curlyC: \curlyX^* \to \curlyX$ is defined by
\begin{equation}
    \curlyC f (g) = \int_\curlyX (f(x) - (f,m))(g(x) - (g,m))\d\P(x).
\end{equation}
We see from the definition that the covariance operator can also be thought of as a bi-linear form on $\curlyX^*$.  

The definition of a Gaussian measure given above allows us to view each $f \in \curlyX^*$ as an element of 
$$L^2(\P) := \left\{f: \curlyX \to \R\; \big\lvert \;\int_\curlyX f(x)^2\d\P(x) < \infty\right\}.$$
Thus, we have an embedding of $\curlyX^* \to L^2(\P)$.  Forming the completion of the image of this embedding with respect to the $L^2(\P)$ norm forms the \emph{Cameron Martin} space of the measure $\P$, denoted $\curlyH_\P$.  Note that by construction, $\curlyX^*$ can be identified with a dense subspace of $\curlyH_\P$ and we have the inclusion map $\mathcal{I}: \curlyH_\P \to L^2(\P)$.  The map $\curlyI$ is sometimes known as the Paley-Wiener map (for example, when $\P$ is the Wiener measure $\curlyI(h)(x)$ is the Ito integral $\int \dot h \d x(t)$).  Note that this implies for any $h \in \curlyH_\P$, the quantity 
$$\langle h, x \rangle^{\sim} := \mathcal{I}(h)(x)$$
is well defined for $\P$-almost-every $x$.  It can additionally be shown that there exists a dense injection $\curlyH_\P \hookrightarrow X$.  The Cameron Martin space has a number of equivalent definitions and determines many of the relevant properties of the measure $\P$.

We will make repeated use of the \emph{Cameron Martin Theorem}, which determines when the translation of a Gaussian measure $\P$ by $h \in X$ gives an equivalent measure.  This will allow us to give an expression for the log likelihood as well as the KL divergence term in the ELBO objective.  For a proof see \cite{bogachev2015gaussian}, \cite{kuo1975gaussian} or \cite{stroock2010probability}.

\begin{theorem}[Cameron-Martin] \label{thm: CM}
Given a Gaussian measure $\P$ on $\curlyX$ with Cameron Martin space $\curlyH_\P$, the translated measure $\P_h(A) := \P(A - h)$ is absolutely continuous with respect to $\P$ if and only if $h \in \curlyH_\P$, with
\begin{equation} \label{cm theorem rn}
    \log \frac{\d \P_h}{\d \P}(x) = -\frac{1}{2}\|h \|_{\curlyH_{\P}}^2 +  \langle h, x \rangle^\sim.
\end{equation}
\end{theorem}

When the Gaussian measure $\P$ is supported on a Hilbert space $\curlyH$, the previous definitions simplify due to the Riesz representation theorem which allows us to use the isomorphism $\curlyH^* \simeq \curlyH$.  In particular, for this case the covariance operator is a trace-class, self-adjoint operator $\curlyC: \curlyH \to \curlyH$.  Further, in this case the Cameron Martin space can be identified with $\text{im}(\curlyC^{1/2})$ and has an inner product given by
$$\langle x, y \rangle_{\curlyH_\P} = \langle \curlyC^{-1/2} x, \curlyC^{-1/2} y \rangle_\curlyH.$$

\subsection{Abstract Wiener Space} \label{sec: AWS}

There is an alternate characterization of Gaussian measures which instead begins with a separable infinite dimensional Hilbert space $\curlyH$.  If we attempt to sample from the ``standard Gaussian" on $\curlyH$ by taking an orthonormal basis $\{e_i\}_{i=1}^\infty$ and try to form the sum
\begin{equation} \label{bad sum}
    x = \sum_{i=1}^\infty \xi_i e_i, \quad \xi_i \sim \mathcal{N}(0,1),\;\text{i.i.d.},
\end{equation}
we see that $\E[\|x\|_\curlyH^2] = \infty$ and thus $x \notin \curlyH$ almost surely.

The way around this is to consider the convergence of the sum \eqref{bad sum} with respect to a norm other than that from $\curlyH$.  After picking such a ``measurable norm" \cite{kuo1975gaussian} we may complete the space $\curlyH$ with respect to this norm to obtain a new Banach space $X$, on which the measure we have tried to construct is supported.  This completion gives us a dense inclusion $i: \curlyH \hookrightarrow X$.  The triple $(i, \curlyH, X)$ is called an \emph{Abstract Wiener Space} (AWS).  The induced Gaussian measure $\P$ on $X$ has $\curlyH$ as its Cameron Martin space.  Thus, the AWS construction gives us a way to construct a Gaussian measure starting from its Cameron Martin space.

An example of this construction that will be particularly useful is that which starts from $L^2(X; \R)$ with $X \subset \R^d$ compact as the desired Cameron Martin space.  If we consider the dense inclusion $L^2(X; \R) \hookrightarrow \curlyH^{-s}(X; \R)$, with $s > d/2$, then $(i, L^2(X; \R), \curlyH^{-s}(X; \R))$ is an AWS and the associated measure is called the \emph{white noise measure}.

\section{Proof of Theorem \ref{thm: fELBO}}

To prove the generalization of the ELBO, we will need to use a modified measure-theoretic formulation of Bayes theorem phrased in terms of Radon-Nikodym derivatives.  We begin from that presented in \cite{ghosal2017fundamentals},

\begin{equation} \label{eq: bayes starting point}
\frac{\d \P_{z|u}}{\d \P_z} = \frac{1}{c(u)}\frac{\d \P_{u|z}}{\d \P_u},
\end{equation}
where
\begin{equation}
c(u) = \int_\curlyZ \frac{\d\P_{u|z}}{\d\P_u}(u)\d \P_z.
\end{equation}
We claim that $c(u) = 1$, $\P_u$-a.e. Since the Radon-Nikodym derivative is non-negative, we have that $c(u) \geq 0$.  Next, we show that $c(u) \leq 1$.  Assume this is not true.  Then by the disintegration property of the regular conditional measures we may write for any measurable $f(u)$,
\begin{align*}
\int_\curlyX f(u) \d\P_u &= \int_\curlyZ \int_\curlyX f(x) \d\P_{u|z} \d\P_z \\
&= \int_{\curlyX} f(u) \left(\int_\curlyZ \frac{\d\P_{u|z}}{\d\P_u} \d\P_z\right)d\P_u \\
&> \int_\curlyX f(u) \d\P_u,
\end{align*}
which is a contradiction.  Hence, $c(u) \leq 1$.  This allows us to write
\begin{align*}
\int_\curlyX  | 1 - c(x)|\;\d \P_u &= \int_\curlyZ  1 - c(x) \;\d \P_u \\
&= \int_\curlyX 1 - \int_\curlyZ \frac{\d\P_{u|z}}{\d\P_u}(u)\d \P_z\d \P_u \\
&= 1 - \int_\curlyZ \int_\curlyX \frac{\d\P_{u|z}}{\d\P_u}(u) \d\P_u \d\P_z \\
&= 1 - \int_\curlyZ \int_\curlyX \d\P_u \d\P_x \\
&= 0,
\end{align*}
and we have shown that $c(x) = 1$ $\P_u$-a.e and thus from \eqref{eq: bayes starting point}
\begin{equation} \label{eq: rn bayes}
\frac{\d \P_{z|u}}{\d \P_z} = \frac{\d \P_{u|z}}{\d \P_u}.
\end{equation}

We are now able to prove the generalize ELBO stated in Theorem \ref{thm: fELBO}.  By the definition of the $\KL$ divergence,
\begin{equation} \label{variational kl}
\mathrm{KL}[\Q^{\phi}_{z|u}||\P_{z|u}] = \int_\curlyZ \log\left(\frac{\d\Q^{\phi}_{z|u}}{\d\P_{z|u}}(z)\right)\;\d\Q^\phi_{z|u}.
\end{equation}
From the chain rule for Radon-Nikodym derivatives we may write
\begin{equation} \label{first rn rewrite}
\frac{\d\Q^\phi_{z|u}}{\d\P_{z|u}} = \frac{\d\Q^\phi_{z|u}}{\d\P_{z}} \frac{\d\P_z}{\d\P_{z|u}}.
\end{equation}
Using \eqref{eq: rn bayes}, we then have
\begin{align}
\frac{\d\Q^\phi_{z|u}}{\d\P_{z|u}} &= \frac{\d\Q^\phi_{z|u}}{\d\P_{z}} \frac{\d\P_u}{\d\P_{u|z}} \nonumber \\
&= \frac{\d\Q^\phi_{z|u}}{\d\P_{z}} \frac{\d\P_u}{\d\V} \frac{\d\V}{\d\P_{u|z}} \nonumber\\
&= \frac{\d\Q^\phi_{z|u}}{\d\P_{z}} \frac{\d\P_u}{\d\V} \left(\frac{\d \P_{u|z}}{\d \V}\right)^{-1},
\end{align}
where the last equality holds under the assumption of mutual absolute continuity between all written measures.  Placing this relation in \eqref{variational kl} shows that
\begin{align}
\mathrm{KL}[\Q^\phi_{z|u}||\P_{z|u}] &= \int_\curlyZ \log\left(\frac{\d\Q^\phi_{z|u}}{\d\P_{z}} \frac{\d\P_u}{\d\V} \left(\frac{\d \P_{u|z}}{\d \V}\right)^{-1}\right)\;\d\Q^\phi_{z|u} \nonumber \\
&= \int_\curlyZ \log\left(\frac{\d\Q^\phi_{z|u}}{\d\P_{z}}\right) \d\Q^\phi_{z|u} +  \int_\curlyZ \log\left(\frac{\d\P_u}{\d\V}\right) \d\Q^\phi_{z|u} - \int_\curlyZ \log \left(\frac{\d \P_{u|z}}{\d \V}\right) \d \Q^\phi_{z|u}.
\end{align}
We identify the first term on the right as $\KL[\Q^{\phi}_{z|u}||\P_z]$.  The integrand of the second term is a function of $u$ only and $\Q^{\phi}_{z|u}$ is a probability measure, thus the second term is equal to the log likelihood of the data $x$.  The third term is the expectation of the conditional likelihood of $x$ given $z$.  We have thus proved the equality
\begin{equation} \label{pre ELBO}
\mathrm{KL}[\Q^\phi_{z|u}||\P_{z|u}] = \KL[\Q^\phi_{z|u}||\P_z] + \log \left(\frac{\d\P_u}{\d\V}(x)\right) - \E_{z \sim \Q^\phi_{z|u}} \left[ \log \frac{\d \P_{u|z}}{\d \V(x)}\right].
\end{equation}

\section{Architectures}

Here we present different architecture choices we have considered in the different experiments presented in this manuscript. Specifically here we outline our specific choices in terms of encoder and decoder architectures, as well as different types of positional encodings.

\subsection{Encoders}
\label{sec:enc choices}
We consider two types of encoders in our benchmarks. In the Gaussian Random Field we consider a Multi-layer Perceptron (MLP) network encoder. In all other benchmarks we build encoders using a simple VGG-style deep convolutional network \cite{simonyan2014very}, where in each layer the input feature maps are down-sampled by a factor of 2 using strided convolutions, while the number of channels are doubled.

\subsection{Decoders}
\label{sec:dec choices}
First, we consider possible decoder choices. The decoders can be categorized broadly as linear and nonlinear. 

\paragraph{Linear Decoder:}

Linear decoders take the form
\begin{equation*} 
\curlyD(z)(x) = \sum_{i=1}^{n} z_i \tau_i(\gamma(x)), \quad z_i\sim \mathcal{N}(0,1), \quad \text{i.i.d.},
\end{equation*}
where $\tau$ is a vector-valued function parameterized by a Multi-layer Perceptron Network (MLP), and $\gamma(x)$ is a positional encoding of the query locations (see next section for more details).

\paragraph{Nonlinear Decoders:}

Generalized nonlinear decoders can be constructed as
\begin{equation*} 
\curlyD(z)(x) = f_\theta(z_i, \gamma(x)), \quad z_i\sim \mathcal{N}(0,1), \quad \text{i.i.d.},
\end{equation*}
where $f$ is a function parameterized by an MLP network, and $\gamma(x)$ is a positional encoding of the query locations (see next section for more details). Following the work of Rebain et. al. \cite{rebain2022attention} we consider two types of conditioning $f$ on the latent vector $z$. The first approach concatenates the latent vector $z$ with the query location $\gamma(x)$ before pushing them through the decoder, as in \cite{seidman2022nomad}. An alternative approach is to split the latent vector into chunks and concatenate each chunk to each  hidden layer of the decoder, see Rebain et. al. for more details.

\subsection{Positional Encodings} 

Positional encodings have been shown to help coordinate-MLPs to capture higher frequency components of signals thereby mitigating spectral bias \cite{mildenhall2021nerf, tancik2020fourier, tancik2020fourier}. Here we employ different types of positional encodings depending on the nature of the benchmark we are considering.

\paragraph{Fourier Features:}
First we consider a periodic encoding of the form
\begin{equation*}
    \gamma(x) = [1, \cos(\omega x), \sin(\omega x), \dots, \cos(k \omega x), \sin(k \omega x)],
\end{equation*}
with $\omega=\frac{2\pi}{L}$, and some non-negative integer $k$. Here $L$ denotes the length of the domain. This encoding was employed in the GRF benchmark to ensure that the generated functions satisfy a zero Dirichlet boundary condition at the domain boundaries.
\paragraph{Random Fourier Features:}
The Random Fourier Feature encoding can be written as:
\begin{equation*}
    \gamma(x) = [\cos(2 \pi B x), \sin(2 \pi B x)],
\end{equation*}
where $B \in \mathbb{R}^{q \times d}$ is sampled from the Gaussian distribution $\mathcal{N}(0, \sigma^2)$ using a user-specified variance parameter $\sigma$. Here $q$ denotes the number of Fourier features used and $d$ is the dimension of the query point $x$. In our experiments we set $q=n/2$, where $n$ is the latent space dimension of the model. We empirically found that this positional encoding gives good performance for the Cahn-Hilliard data-set. 
\paragraph{Multi-resolution Hash Encoding:} 
For cases where we expect the function to contain multiscale characteristics, such as in the InSAR data-set, we considered the Multi-resolution Hash Encoding proposed in \cite{muller2022instant}. We found that this type of encoding gives the best performance for the InSar data-set. 

\subsection{Hyper-parameter Sweeps} 

In order to quantify the sensitivity of our results on the choice of different hyper-parameters, as well as to identify hyper-parameter settings that lead to good performance, we have performed a series of  hyper-parameter sweeps for each benchmark considered in the main manuscript. Here we present the sweep settings and the best configuration identified for each benchmark.

\paragraph{Gaussian Random Field:} We perform a parametric sweep by considering different latent dimension sizes and different model initialization seeds, as shown in Table \ref{tab:sweep_choices}. Our goal is to find the model out of these set-ups that minimizes the normalized Hilbert-Schmidt norm between samples drawn from the ground truth data distribution and the VANO models. The lowest Hilbert-Schmidt norm value is achieved for $n=64$.

\paragraph{2D Gaussian Densities:} We perform a parametric sweep by considering different latent dimension sizes and different model initialization seeds, as shown in Table \ref{tab:sweep_choices}. Our goal is to find the model out of these set-ups that minimizes the generalized MMD metric between samples drawn from the ground truth data distribution and the VANO models. The lowest generalized MMD value is achieved for $n=32$.

\paragraph{Cahn-Hilliard:} We perform a parametric sweep by considering different latent dimension sizes,  KL loss weights $\beta$ and decoder types, as shown in Table \ref{tab:sweep_choices}. Our goal is to find the model out of these set-ups that minimize the generalized MMD metric between samples drawn from the ground truth data distribution and the VANO models. The lowest generalized MMD value is achieved for $n=64$, $\beta=10^{-4}$ and a concatenation decoder.

\paragraph{InSAR Interferograms:} We perform a parametric sweep by considering different latent dimension sizes,  KL loss weight $\beta$ and decoder layer width, as shown in Table \ref{tab:sweep_choices}. Our goal is to find the model out of these set-ups that minimize the generalized MMD metric between samples drawn from the ground truth data distribution and the VANO models. The lowest generalized MMD value is achieved for $n=256$, $\beta=10^{-4}$ and a decoder layer width of $512$.

\begin{table}
\caption{Hyper-parameter sweep settings for different examples (C and SC indicate a concatenation and a split concatenation decoder, respectively, see Section \ref{sec:dec choices}).}
\label{tab:sweep_choices}
\vskip 0.15in
\begin{center}
\begin{small}
\begin{sc}
\begin{tabular}{lccccr}
\toprule
Benchmark & random seed & $n$ & $\beta$ & decoder & layer width \\ 
\midrule
 GRF & [0, \ldots, 10] & [2,4,8,16,32,64]  & - & - & -\\
\midrule
 2D Gaussian Densities & [0, \ldots, 4] & [4, 32, 64, 128, 256]  & - & - & -\\
\midrule
 Cahn-Hilliard  & - & [32, 64, 128] & [$10^{-5}$, $10^{-4}$, $10^{-3}$] & [C,SC] & [64, 128, 256]\\
\midrule
 InSAR Interferograms &  - & [128, 256, 512] & [$10^{-5}$, $10^{-4}$, $10^{-3}$] & - & [256, 512] \\
\bottomrule
\end{tabular}
\end{sc}
\end{small}
\end{center}
\vskip -0.1in
\end{table}

\section{Experimental Details}
\label{sec:exp details}
In this section we present details about each experimental set-up and the generation of the training and testing data-sets. 

\subsection{Gaussian Random Field}

\paragraph{Data Generation:}

From the Karhunen-Lo\'eve theorem \citep{adler1990introduction} we can construct random functions distributed according to a mean zero Gaussian measure with covariance $\Gamma$
\begin{align}
u = \sum_{i=1}^\infty \xi_i \sqrt{\lambda_i} \phi_i,
\end{align}
where $\xi_i \sim \mathcal{N}(0,1)$ are i.i.d. normally distributed random variables, and $\lambda_i$ and $\phi_i$ are the eigenvalues and eigenvectors of the covariance operator.

We define $\lambda_i = ((2 \pi i)^2  + \tau^2)^{-\alpha}$ and $\phi_i(x)= \sqrt{2} \sin(2 \pi i x)$.  These setup corresponds to the eigenvalues and eigenfunctions of the operator $(I + \Delta)^{-\alpha}$ on $X = [0,1]$ with zero boundary conditions. To generate these functions, we take samples of the sum in \eqref{eq: KL mu} truncated at the first $32$ eigenfunctions and eigenvalues to construct $N_{train} = 2048$ functions evaluated at $m = 128$ points which we use to train our model.  We independently sample an additional $N_{test} = 2048$ functions for testing.

\paragraph{Encoder:}
We parameterize the encoder using an MLP network with 3 hidden  layers, width of 128 neurons and Gelu \cite{hendrycks2016gaussian} activation functions. 
 
\paragraph{Decoder:}
We consider a linear decoder parameterized by a 3-layer deep MLP network, with a width of 128 neurons width, periodic positional encoding and Gelu  activation functions.

\paragraph{Training Details:}
We consider a latent space dimension of $n=64$, $S=16$ Monte Carlo samples for evaluating the expectation in the reconstruction loss, and a KL loss weighting factor $\beta=5 \ 10^{-6}$. We train the model using the Adam optimizer \cite{kingma2014adam} with random weight factorization \cite{wang2022random} for  $40,000$ training iterations with a batch size of 32 and a starting learning rate of $10^{-3}$ with  exponential decay of 0.9 every 1,000 steps. 

Additionally, it has been observed in the operator learning literature \cite{di2023neural, wang2022improved} that when a functional data-set has examples of varying magnitudes the training can overfit those with larger magnitude, as they are more heavily weighted in the loss function.  To correct for this, we use a data dependent scaling factor in the likelihood terms.  We find this modification is only necessary for the first two experiments, as the other scenarios do not have this magnitude variability in their data.

\paragraph{Evaluation:}
To test the performance of the linear decoder VANO model, we train it on a range of dimensions for its latent space, $n \in \{ 2, 4, 8, 16, 32, 64\}$.  We then generate data by sampling  latent vectors from the prior $z \sim \P_z = \mathcal N(0,\mathrm{I})$ and map them through the decoder to create functions as in \eqref{eq: linear gen}.  Since this distribution is a mean zero Gaussian, like the ground truth Gaussian random field, it suffices to compare the resulting covariance operators to measure how close the two distributions are.  The covariance associated with the distribution described in \eqref{eq: linear gen} is given by 
\begin{equation}
\hat \Gamma = \sum_{i=1}^n \tau_i \otimes \tau_i.
\end{equation}
We use the ground truth covariance eigenfunctions and the approximated eigenfunctions to compute the true and approximated covariance operators, and compare them using the Hilber-Schmidt norm. We observe in Figure \ref{fig:grf1d_sweep_boxplot} that qualitatively the two sets of look similar indicating the model is recovering the optimal choice of decoder.

\begin{figure}
\begin{center}
\centerline{\includegraphics[width=0.75\columnwidth]{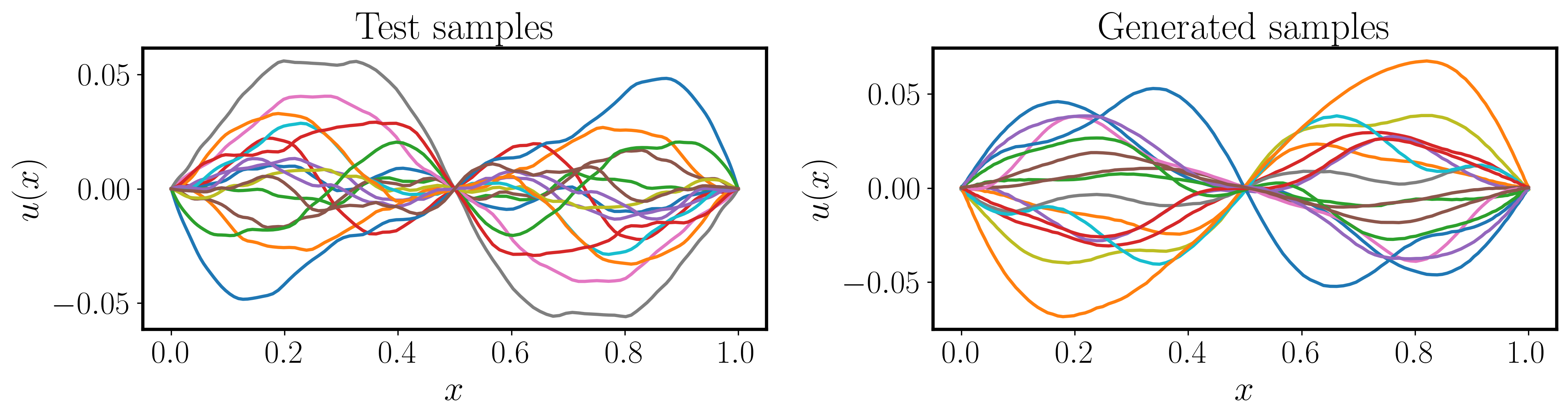}}
\caption{{\it GRF benchmark:} Left: Functions sampled from the ground truth datset, Right: Functions sampled from the VANO model.}
\label{fig:generated_samples_comparison}
\end{center}
\end{figure}

\subsection{2D Gaussian Densities}

\paragraph{Data Generation:} For this example we construct a functional data-set consisting of two-dimensional Gaussian pdf functions in the unit square $X=[0,1]^2$

\begin{equation}
\begin{aligned}
        U(x, y) = (2 \pi)^{-1/2} \det (\Sigma)^{-1/2} 
         \exp \Big ( - \frac{1}{2} (x - \mu)^\top \Sigma^{-1} (x - \mu) \Big), 
\end{aligned}
\end{equation}
where $\mu \sim  \begin{pmatrix} \mu_x, \mu_y \end{pmatrix}$ and $\Sigma = \sigma I$. We consider $\mu_x, \mu_y \sim U(0,1)$ and $\sigma \sim U(0,0.1)+0.01$ and generate $N_{train}=2,048$ training and $N_{test}=2048$ testing samples of $m=2,304$ measurement points on a $48 \times 48$ equi-spaced grid. 

\paragraph{Encoder:}
We parameterize the encoder using a 4-layer deep VGG-style  convolutional network using a $2\times 2$ convolution kernel, a stride of size two, (8, 16, 32, 64) channels at each layer, and Gelu activation functions. 
 
\paragraph{Decoder:}
We consider two types of decoders, a linear and a nonlinear. The linear decoder is parameterized using a 3-layer deep MLP network with 128 neurons per layer and Gelu activation functions. For the nonlinear decoder case we consider a 3-layer deep MLP network with 128 neurons per layers, Gelu activation functions and concatenation conditioning. In both cases we also apply a softplus activation on the decoder output  to ensure positivity in the predicted function values.

\paragraph{Training Details:} 
We consider a latent space dimension of $n=32$, $S=4$ Monte Carlo samples for evaluating the expectation of the reconstruction part of the loss and a KL loss weighting factor $\beta= 10^{-5}$. We train the model using the Adam optimizer \cite{kingma2014adam} with random weight factorization \cite{wang2022random} for  $20,000$ training iterations with a batch size of 32 and a starting learning rate of $10^{-3}$ with  exponential decay of 0.9 every 1,000 steps. For this case, we also re-scale the likelihood terms by the empirical norm of the input function to ensure the terms with larger magnitude do not dominate the optimization procedure, as in \cite{wang2022improved, di2023neural}.

\begin{figure}
\begin{center}
\centerline{\includegraphics[width=0.35\columnwidth]{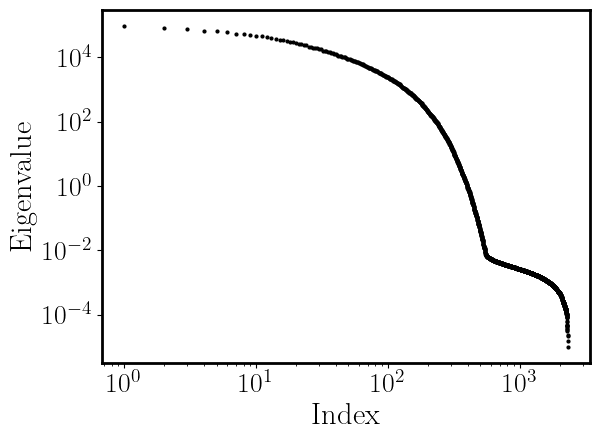}}
\caption{{\it 2D Gaussian densities benchmark:} PCA eigenvalue decay of the training data-set.}
\label{fig:GaussianBump/evalues_decay_gaussian}
\end{center}
\end{figure}

\begin{figure}
\begin{center}
\centerline{\includegraphics[width=\textwidth]{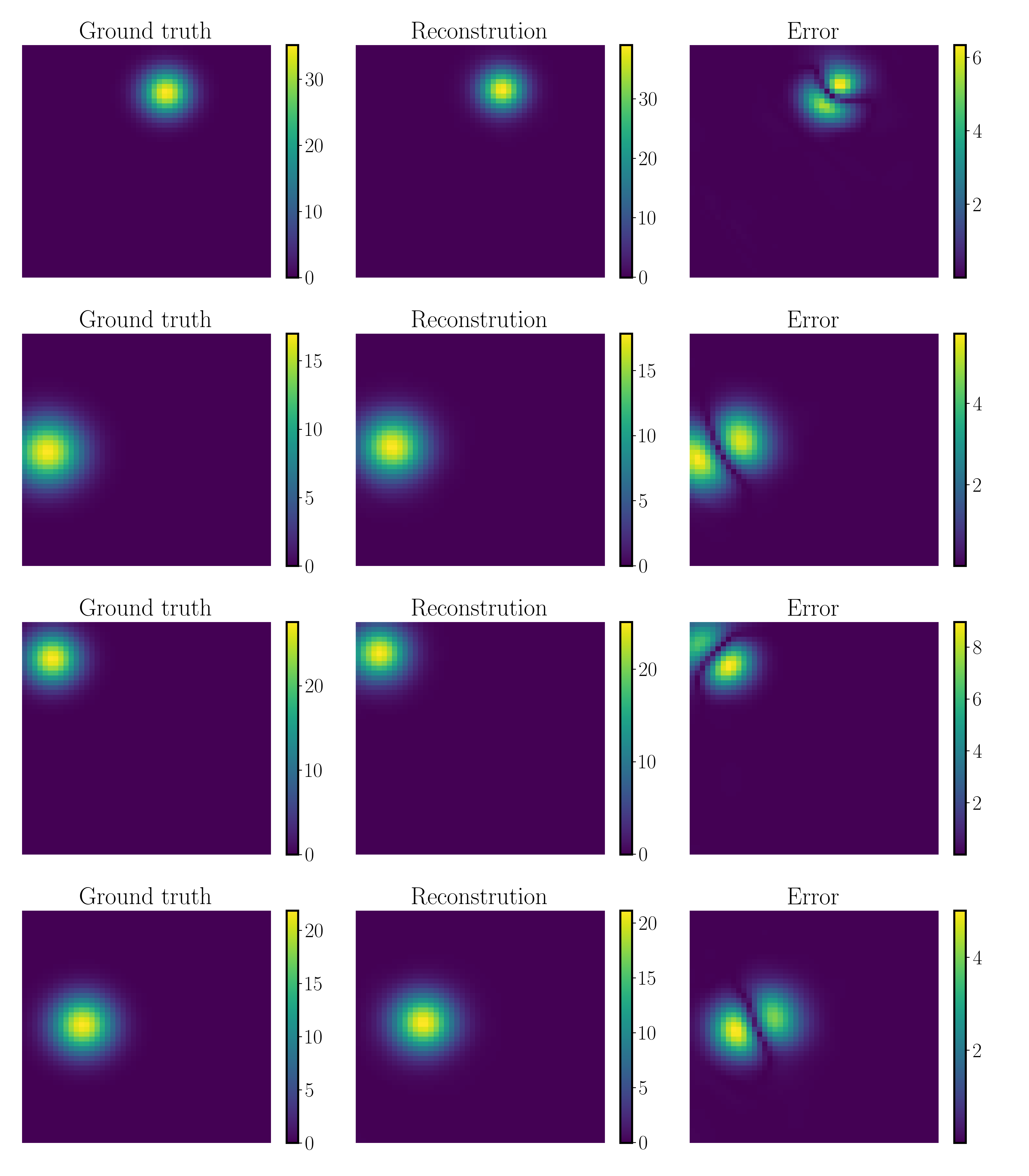}}
\caption{{\it 2D Gaussian densities benchmark:} Left: Ground truth functions samples from the test data-set, Middle: Nonlinear VANO reconstruction, Right: Absolute point-wise error.}
\label{fig:GaussianBump/test_reconstructions}
\end{center}
\end{figure}

\begin{figure}
\begin{center}
\centerline{\includegraphics[width=\textwidth]{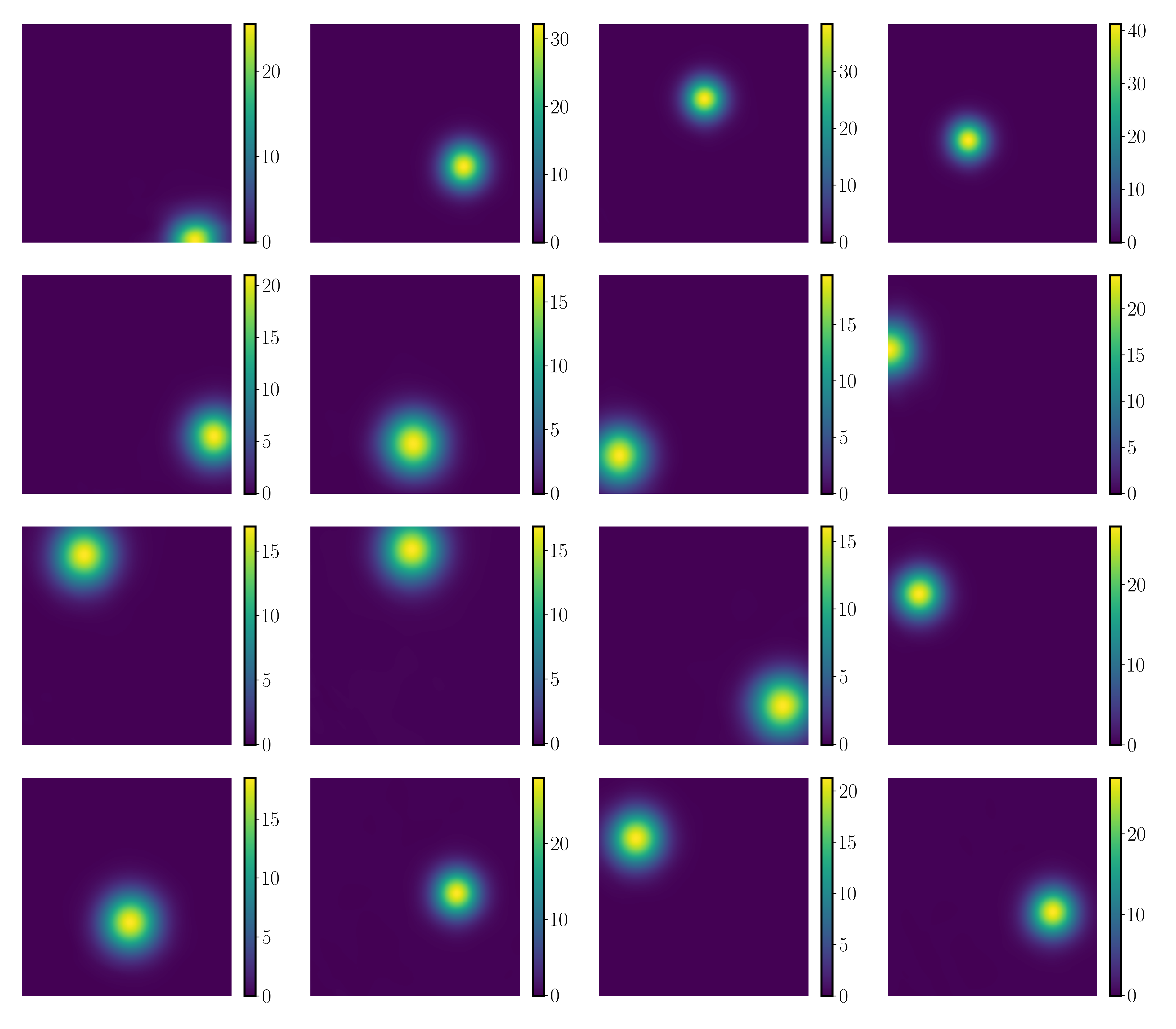}}
\caption{{\it 2D Gaussian densities benchmark:} Gaussian density function samples generated from the  VANO model with a nonlinear decoder in super-resolution mode (training resolution: $48\times 48$, sample resolution $256\times 256$).}
\label{fig:GaussianBump/gaussian_density_samples}
\end{center}
\end{figure}

\paragraph{Evaluation:} In figure \ref{fig:GaussianBump/evalues_decay_gaussian} we present the decay of the PCA eigenvalues computed across function samples. The eigenvalue decay is slow which is the reason that makes the VANO with a linear decoder fail in reconstructing and generating function samples as shown in Figures \ref{fig:GaussianBump/test_reconstructions} and \ref{fig:GaussianBump/gaussian_density_samples}, respectively. We perform a comparison between samples from the true data-set and samples generated by VANO using the generalized MMD metric computed using 512 function samples. 

\begin{figure}
\begin{center}
\centerline{\includegraphics[width=\textwidth]{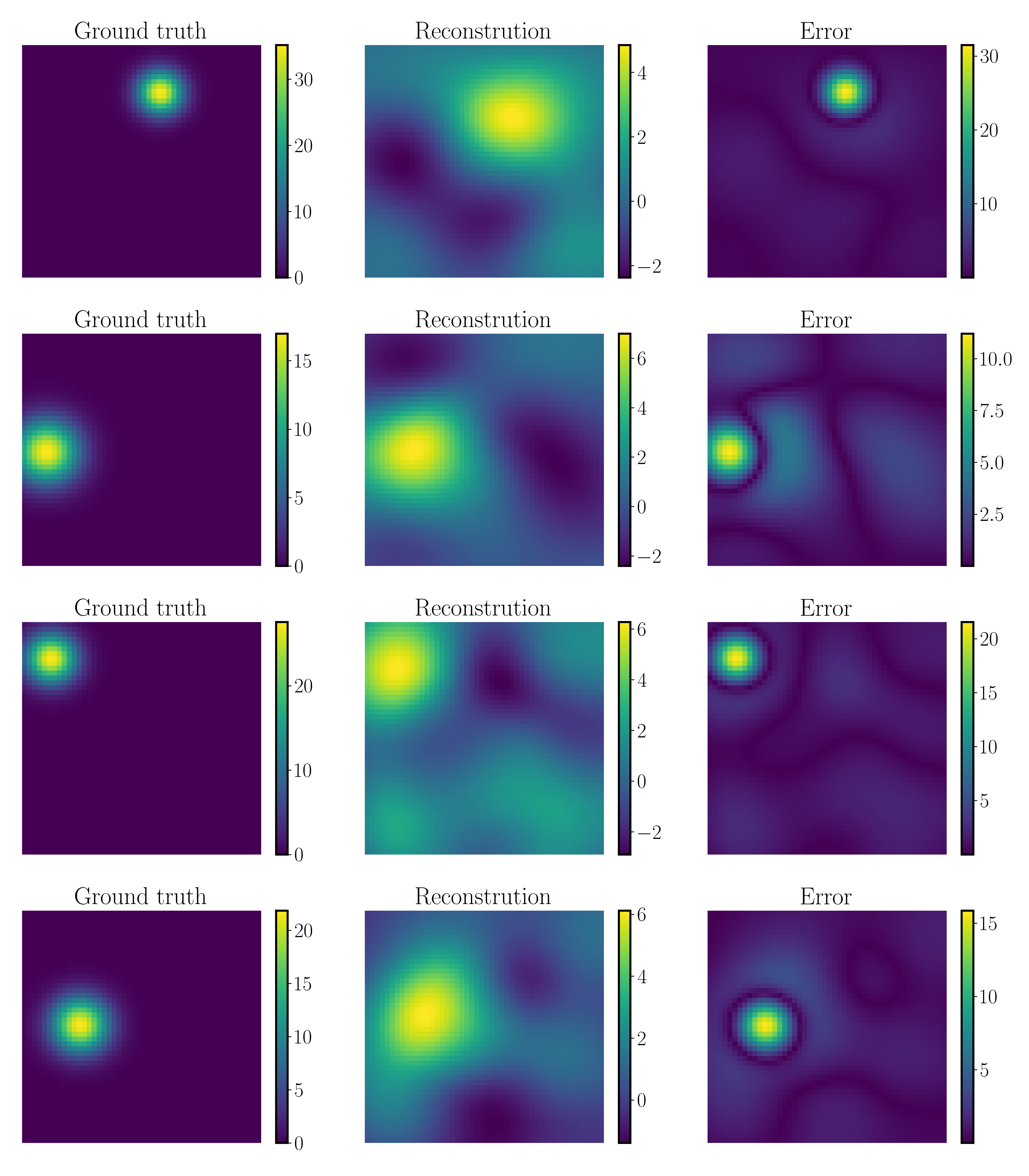}}
\caption{{\it 2D Gaussian densities benchmark:} Left: Ground truth functions samples from the test data-set, Middle: Linear VANO reconstruction, Right: Absolute error.}
\label{fig:GaussianBump/test_reconstruction_linear}
\end{center}
\end{figure}

\begin{figure}
\begin{center}
\centerline{\includegraphics[width=\textwidth]{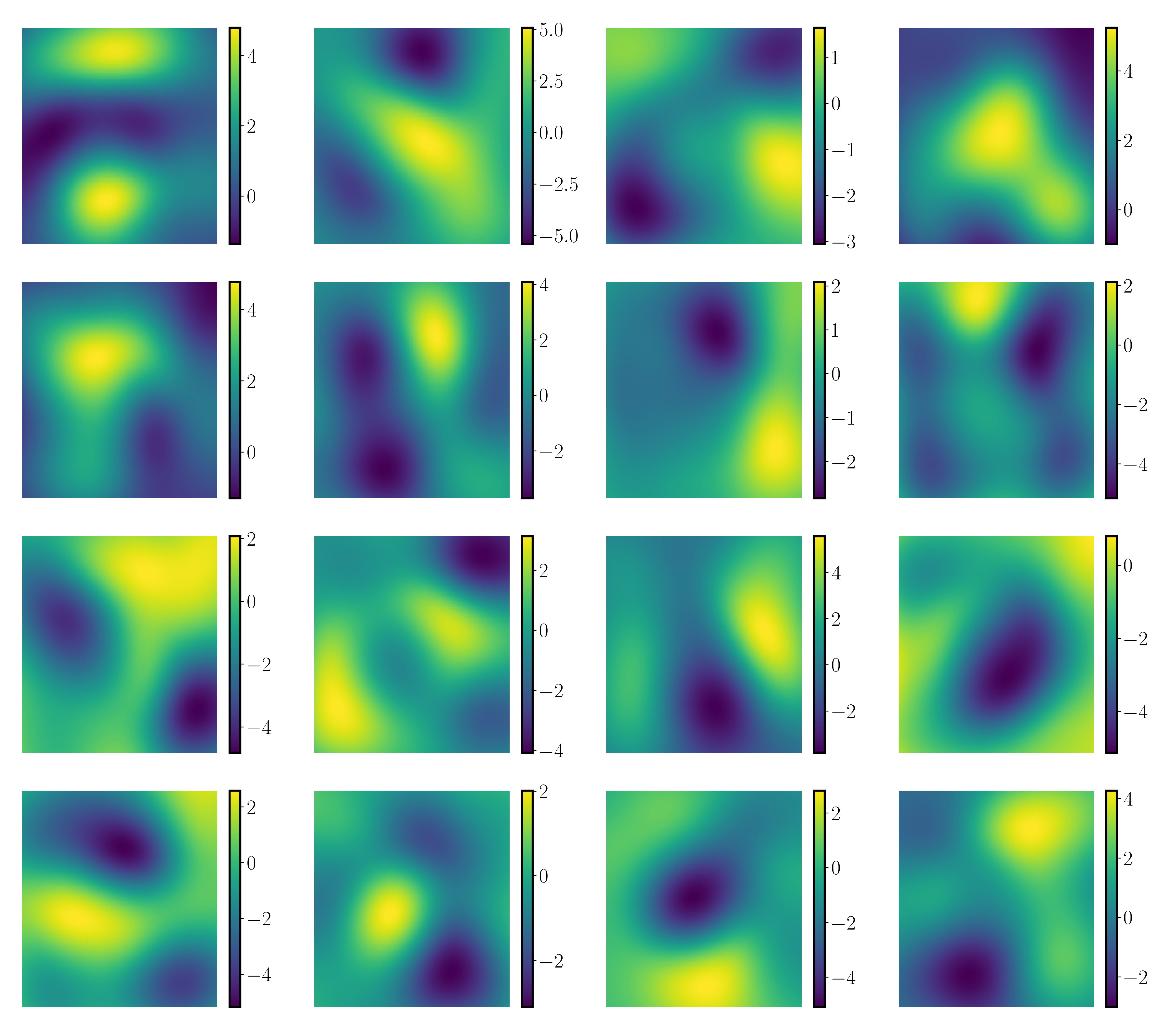}}
\caption{{\it 2D Gaussian densities benchmark:} 2D Gaussian density function samples generated from the  VANO model using a linear decoder in super-resolution mode (training resolution: $48\times 48$, sample resolution $256\times 256$).}
\label{fig:GaussianBump/samples_gaussian_linear}
\end{center}
\end{figure}

\subsection{Phase separation patterns in Cahn-Hilliard systems}
\label{sec:Cahn-Hilliard Appendix}

\paragraph{Data Generation:} The Cahn-Hilliard equation is a fourth order partial differential equation that describes the evolution of the phase separation process in an anisotropic binary alloys \cite{kobeissi2022mechanical}. The Cahn-Hilliard equation is written as:

\begin{equation}
        \frac{\partial c }{\partial t } - \nabla \cdot ( M \nabla ( \mu_c - \lambda \nabla^2 c)) = 0 \quad \text{in} \Omega, 
\end{equation}
where $0<c<1$ denotes the concentration of one of the components of the binary mixture, $M$ is a mobility parameter, $\mu_c$ is the chemical potential of the uniform solution, and $\lambda$ a positive scalar that describes the thickness of the interfaces of the two mixtures. We consider boundary conditions
\begin{equation}
\begin{aligned}
 c &=g \quad \text{on} \ \ \Gamma_g, \\
 M \lambda \nabla c \cdot \bm{n} &= 0 \quad \text{in} \ \ \partial \Omega, \\
 M \lambda \nabla c &= r \quad \text{on} \ \ \Gamma_r, \\
 M \nabla (\mu_c - \lambda \nabla^2 c) \cdot \mathbf{n} &= s \quad \text{on} \ \ \Gamma_s,
\end{aligned}
\end{equation}
where $\Omega$ is two dimensional domain, $\partial \Omega$ the domain boundary, $\bm{n}$ the unit outward normal, and $\overline{\Gamma_g \bigcup \Gamma_s} = \overline{\Gamma_q \bigcup \Gamma_r}$, see \cite{kobeissi2022mechanical}. The initial conditions are given by
\begin{equation}
    c(\mathbf{x}, 0) = c_0(\mathbf{x}) \quad \text{in} \ \ \Omega.
\end{equation}
For the above set-up the chemical potential $\mu_c$ is chosen as a symmetric double well potential where the wells correspond to the two different material phases, namely $f = b c^2 ( 1 - c^2)$. For more information on the set-up we refer the interested reader to \cite{kobeissi2022mechanical}.

\paragraph{Encoder:}
We employ a VGG-style convolutional encoder with 5 layers using $2\times 2$ convolution kernels, stride of size two, (8, 16, 32, 64, 128) channels per layer, and Gelu activation functions. 
 
\paragraph{Decoder:}
We employ a nonlinear decoder parameterized by a 4-layer deep MLP network with 256 neurons per layer, random Fourier Features positional encoding \cite{tancik2020fourier} with $\sigma^2 = 10$, Gelu  activation functions and concatenation conditioning. We also apply a sigmoid activation on the decoder outputs to ensure that the predicted function values are in [0,1].

\paragraph{Training Details:} 
We consider a latent space dimension of $n=64$, $S=4$ Monte Carlo samples for evaluating the expectation of the reconstruction part of the loss and a KL loss weighting factor $\beta= 10^{-4}$. We train the model using the Adam optimizer \cite{kingma2014adam} with random weight factorization \cite{wang2022random} for  $20,000$ training iterations with a batch size of 16 and a starting learning rate of $10^{-3}$ with  exponential decay of 0.9 every 1,000 steps. 

\paragraph{Discretize-first VAE training set-up:} For the discretize-first VAE simulations we consider an identical encoder as in the VANO case, and a convolutional decoder that exactly mirrors the encoder structure, albeit using transposed convolutions. 

\begin{figure}
\begin{center}
\centerline{\includegraphics[width=\textwidth]{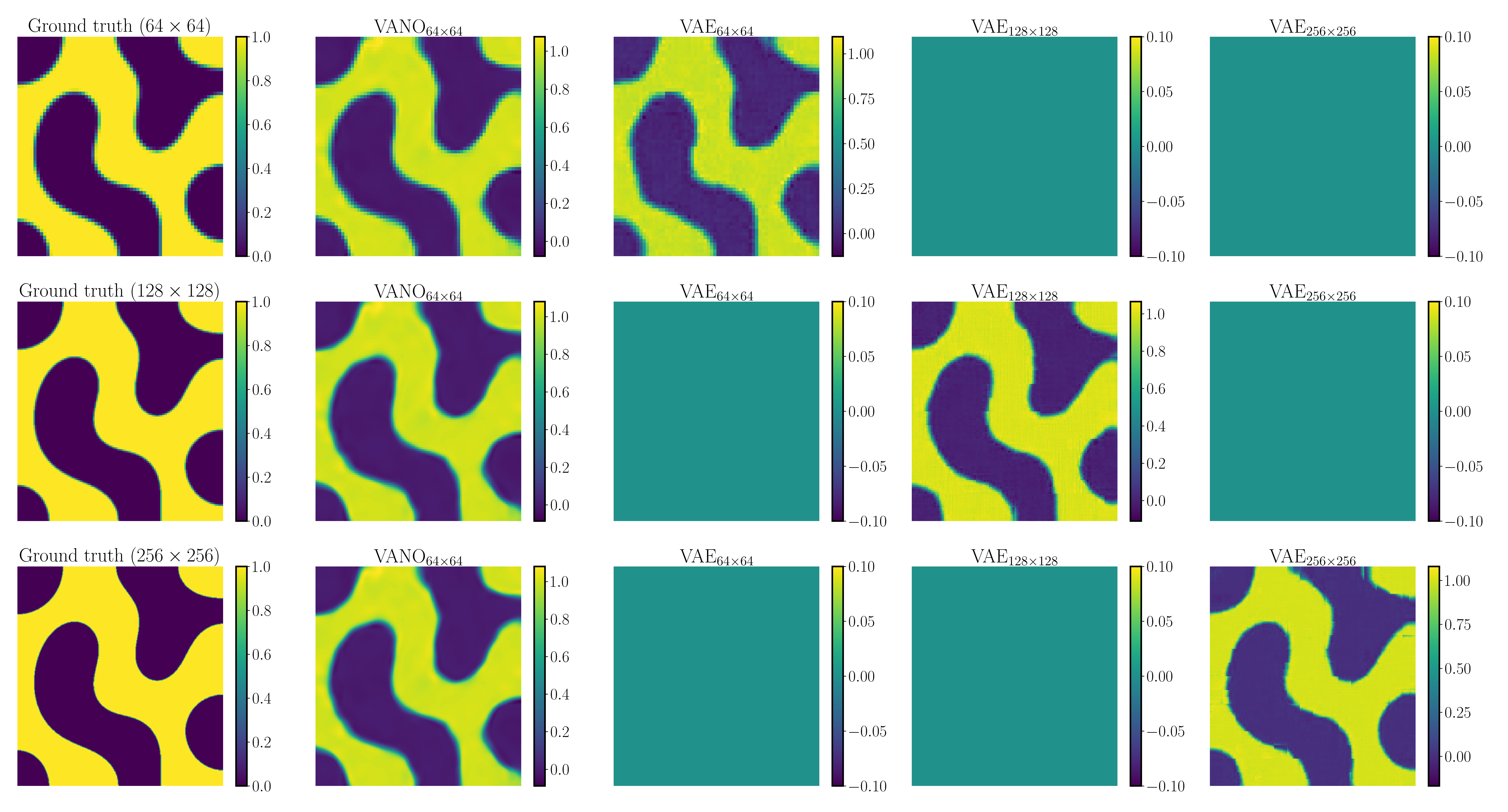}}
\caption{{\it Cahn-Hilliard benchmark:} Ground truth data-set samples, VANO reconstructions, and discretize first VAE reconstructions of Cahn-Hilliard functions in different resolutions. For the VANO model, we train on 64x64 resolution and then perform super-resolution for 128x128 and 256x256, while for the discrete-first VAE models we must train a separate model at each target resolution.}
\label{fig:CahnHilliard/test_reconstructions_comparisons}
\end{center}
\end{figure}

\begin{figure}
\begin{center}
\centerline{\includegraphics[width=\textwidth]{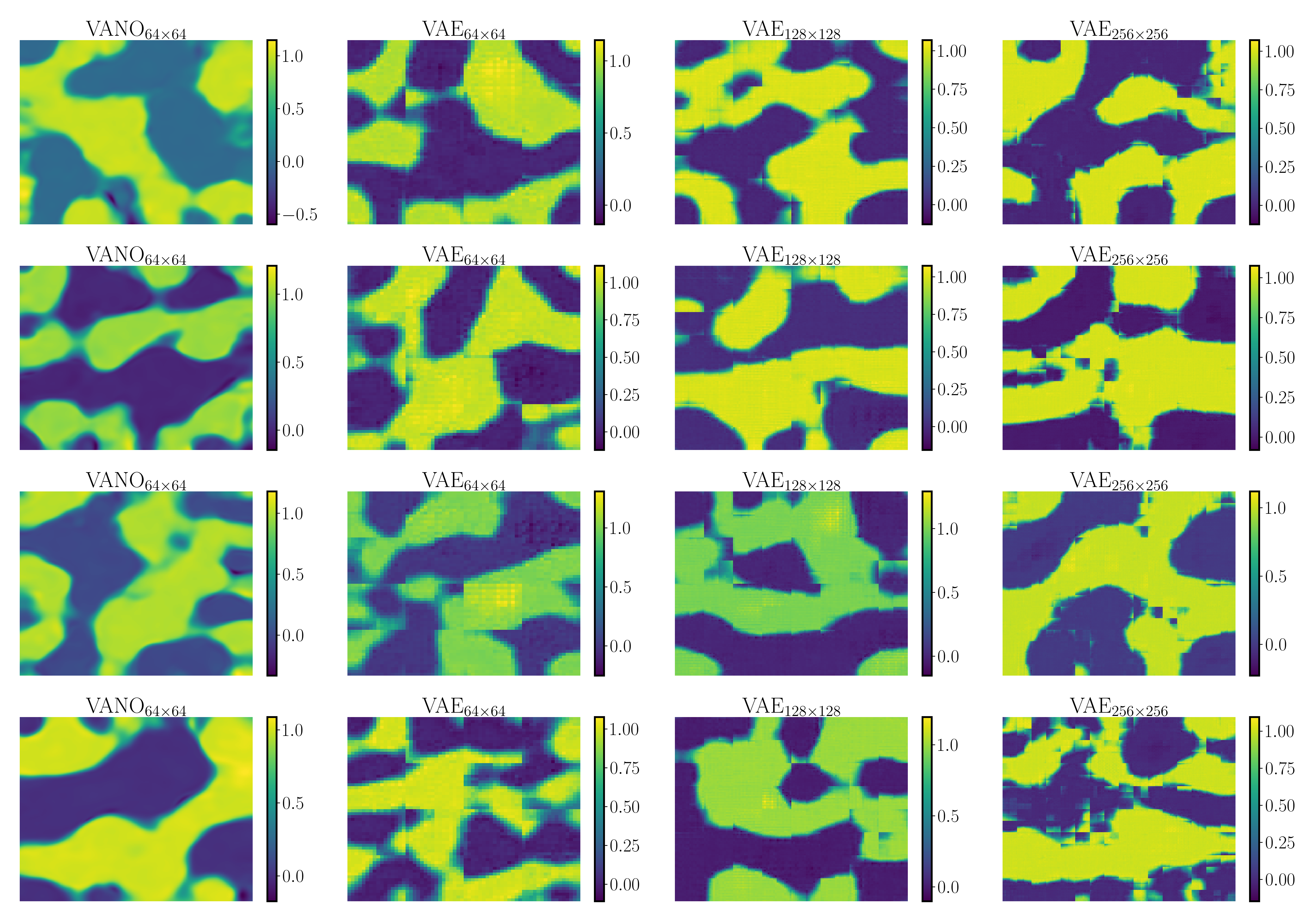}}
\caption{{\it Cahn-Hilliard benchmark:} On the far left column we provide 256x256 image samples from the VANO model trained on 64x64 resolution. On the other columns, we provide discretize-first VAE samples where the training resolution is indicated by their subscript, i.e. VAE$_{64 \times 64}$ indicates a samples coming from a VAE model trained on 64x64 resolution.}
\label{fig:CahnHilliard/samples_comparisons_2}
\end{center}
\end{figure}

\paragraph{Evaluation:}
We perform a comparison between the ground truth test data-set and samples generated by VANO and the discretize-first VAE models using the generalized MMD metric computed using 256 function samples. We present reconstructed function samples chosen randomly from the test data-set for the VANO model trained on a 64x64 resolution and use to make predictions on higher resolutions, namely 128x128 and 256x256. Separate discretize-first VAE models are trained on data with resolutions 64x64, 128x128 and 256x256. Representative test reconstructions from all models are presented in Figure \ref{fig:CahnHilliard/test_reconstructions_comparisons}. In Figure \ref{fig:CahnHilliard/samples_comparisons_2} we show  generated function samples at different resolutions. For the VANO model we train the model on 64x64 resolution images and generate samples in super-resolution mode, while for the discretize-VAE models we present samples of the same resolution as the images on which each model was trained on. The resolution of the images used for training each model are indicated by subscripts.

\subsection{Interferometric Synthetic Aperture Radar data-set}
\label{sec:InSAR Appendix}

\paragraph{Data Generation:} InSAR is a sensing technology that exploits radar signals from aerial vehicles in order to record changes in echoes over time to measure the deformation of a point on the Earth between each pass of the aerial vehicle over the specified point. This technology is employed in measuring deformation on the Earth's surface caused by earthquakes, volcanic eruptions, etc. The data returned by InSAR mostly consists of interferograms.   An interferogram is an angular-valued spatial field $u \in \mathcal{X}$ and $u(x) \in [-\pi, \pi]$ where $x \in X$ the domain of $u$ which in this case corresponds to the Earth surface. Interferograms contain different types of noise and are affected by local weather, Earth topography, and different passes of the aerial vehicles which makes them complex to approximate \cite{rahman2022generative}. 

The InSAR data-set we use consists of $N=4,096$ examples extracted from raw interferograms, each of $128 \times 128$ resolution coming from the Sentinel-1 satellites, as described in \cite{rahman2022generative}. The satellite image covers a 250 by 160 km area around the Long Valley Caldera, an active volcano in Mammoth Lakes, California, from November 2014 to March 2022, using the InSAR Scientific Computing Environment \cite{rosen2012insar}. The data is pre-processed as described in \cite{rahman2022generative} to produce the training data-set. We train VANO on the entire data, as in \cite{rahman2022generative}. 

\paragraph{Encoder:}
We employ a simple VGG-style convolutional encoder with 6 layers using $2\times 2$ convolution kernels, stride of size two, (8, 16, 32, 64, 128, 256) channels per layer, and Gelu activation functions.

\paragraph{Decoder:}
We employ a nonlinear decoder parameterized by an 8-layer deep MLP network with 512 neurons per layer and split concatenation conditioning. We also employ the Multi-resolution Hash Encoding put forth by \cite{muller2022instant} in order to capture the multi-resolution structure in the target functional signals (see \cite{muller2022instant} for the default hyper-parameter settings).

\paragraph{Training Details:} 
We consider a latent space dimension of $n=256$, $S=4$ Monte Carlo samples for evaluating the expectation of the reconstruction part of the loss and a KL loss weighting factor $\beta= 10^{-4}$. We train the model using the Adam optimizer \cite{kingma2014adam} with random weight factorization \cite{wang2022random} for  $20,000$ training iterations with a batch size of 16 and a starting learning rate of $10^{-3}$ with  exponential decay of 0.9 every 1,000 steps. 

\paragraph{GANO training setup:} We use the implementation from the official repository of the GANO paper \footnote{\url{https://github.com/kazizzad/GANO}} to train the model with the recommended hyper-parameter settings. 

\begin{figure}
\begin{center}
\centerline{\includegraphics[width=\textwidth]{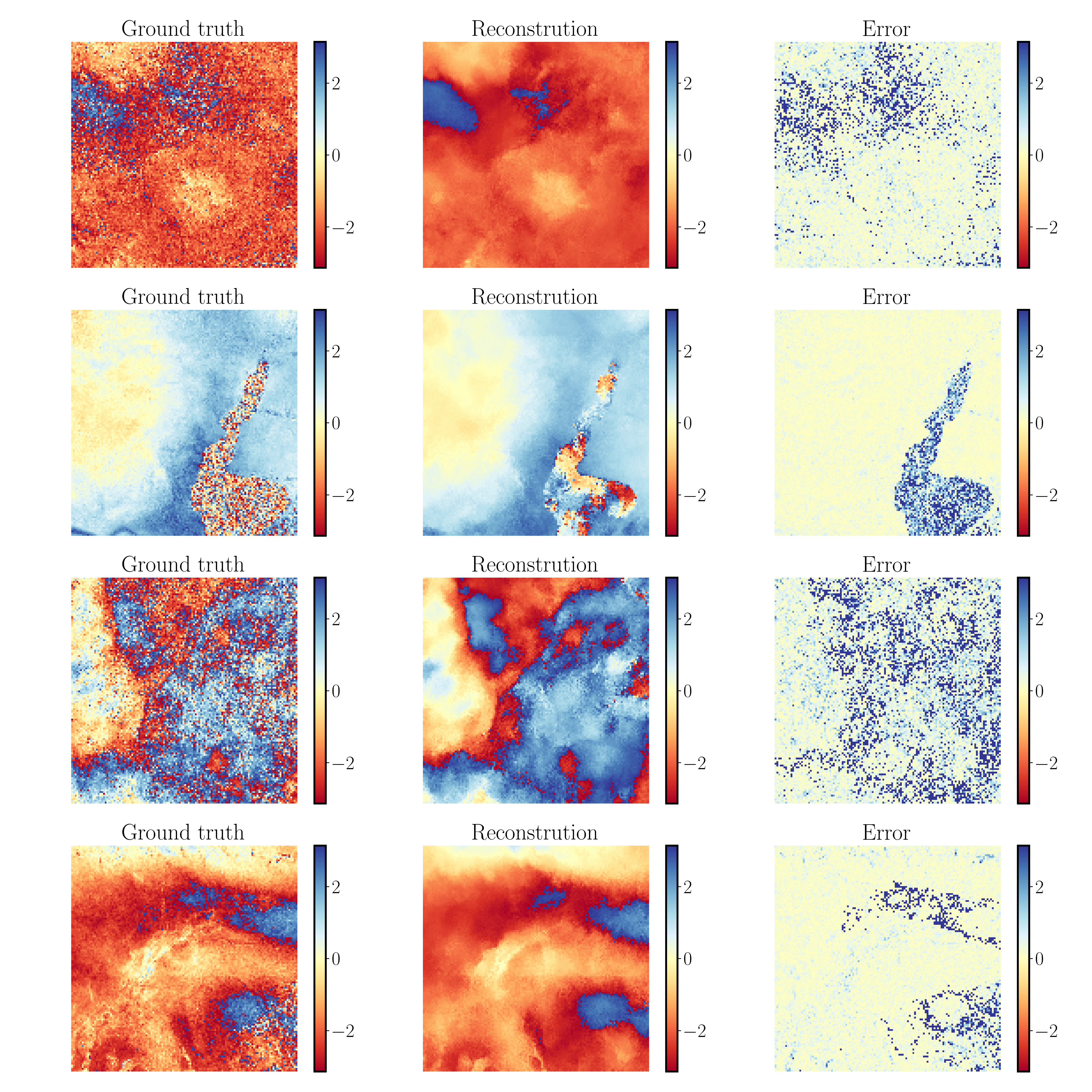}}
\caption{{\it InSar benchmark:} Left: Ground truth functions samples from the data-set, Middle: Linear VANO reconstruction, Right: Absolute error.}
\label{fig:Volcano/test_reconstructions}
\end{center}
\end{figure}

\begin{figure}
\begin{center}
\centerline{\includegraphics[width=\textwidth]{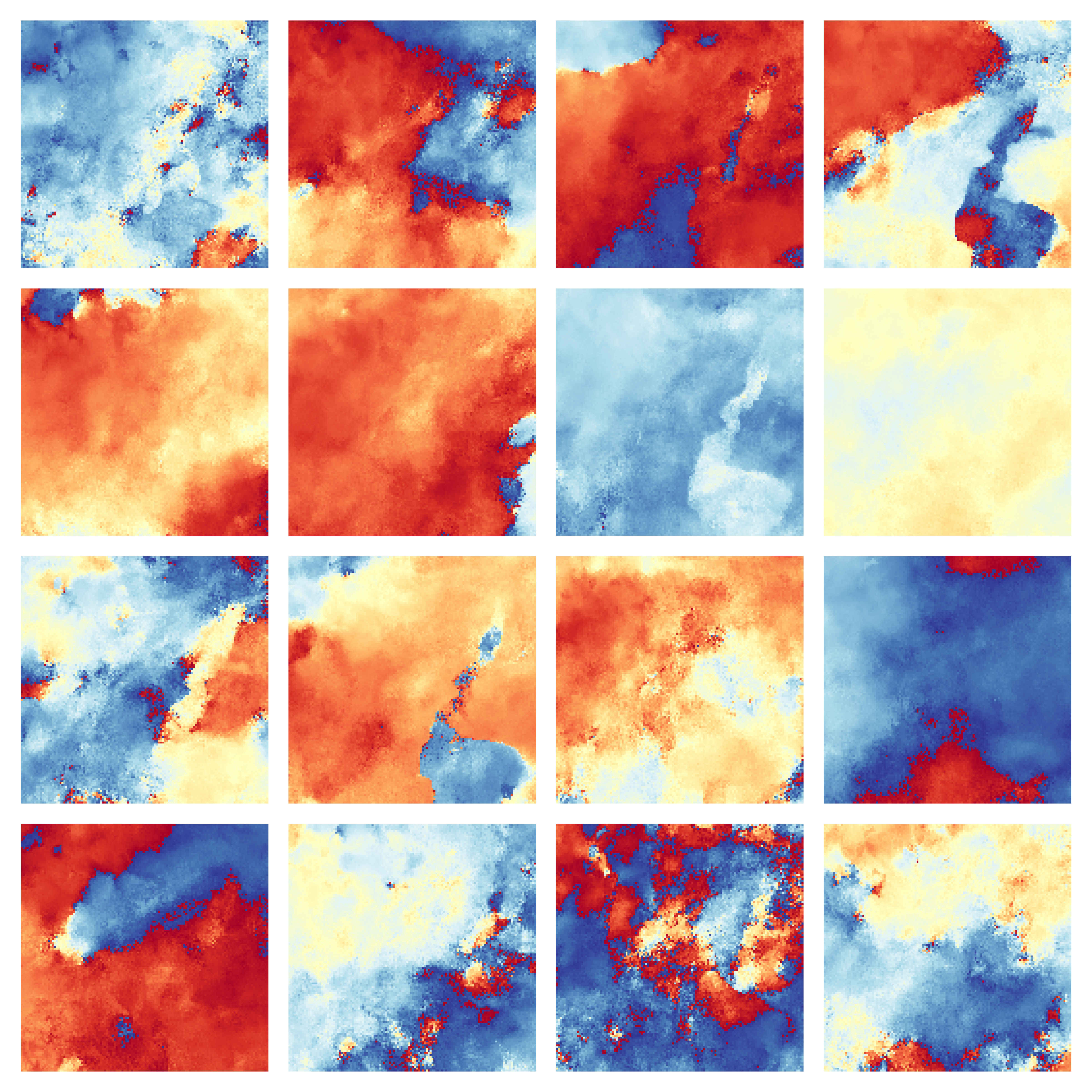}}
\caption{{\it InSar benchmark:} VANO generated function samples for InSAR Interferograms.}
\label{fig:Volcano/vano_samples}
\end{center}
\end{figure}

\begin{figure}
\begin{center}
\centerline{\includegraphics[width=\textwidth]{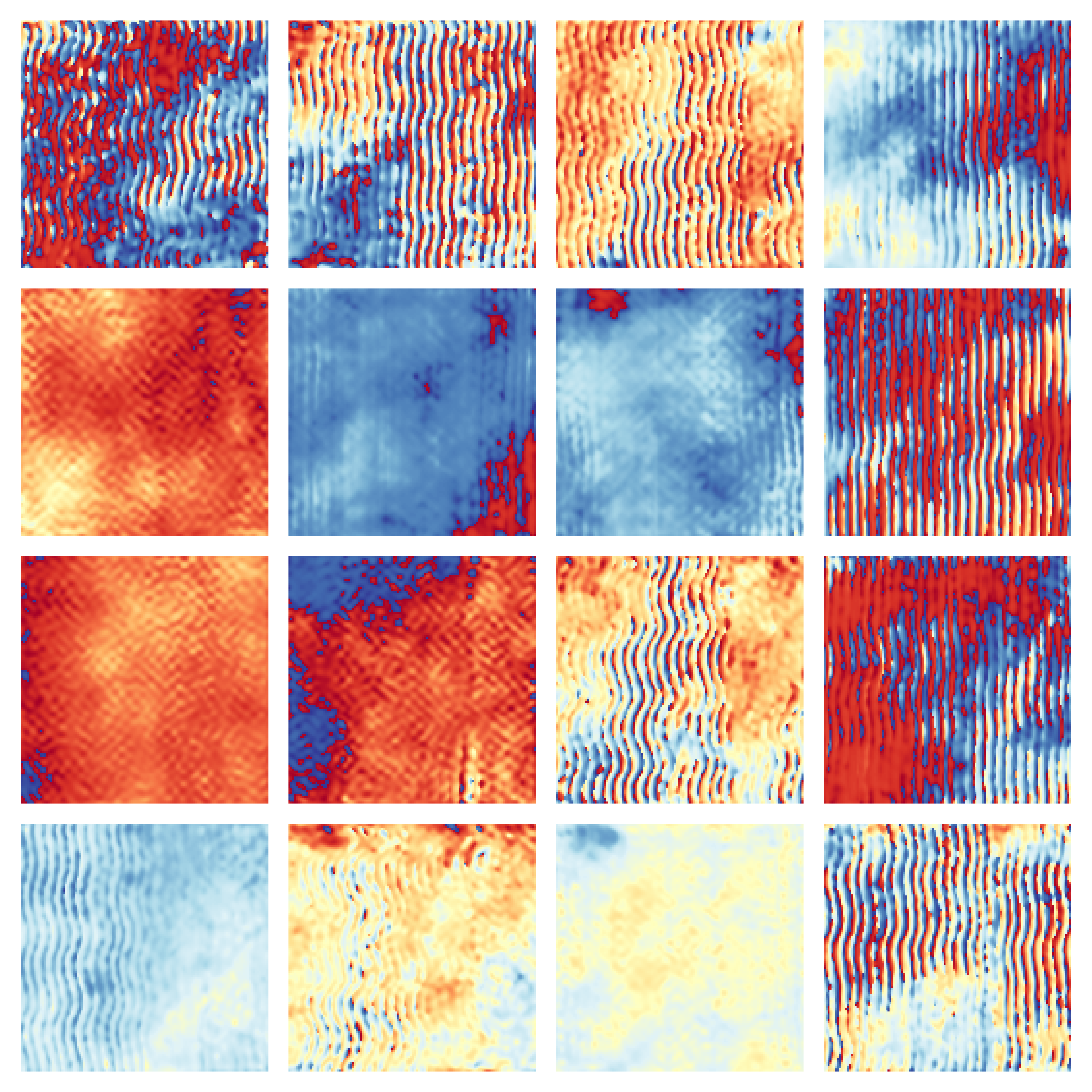}}
\caption{{\it InSar benchmark:} GANO \cite{rahman2022generative} generated function samples for InSAR Interferograms.}
\label{fig:Volcano/gano_samples}
\end{center}
\end{figure}

\paragraph{Evaluation:}
We evaluate the performance of our model using two metrics: the circular variance and  the circular skewness, as explained in the Appendix Section \ref{sec: circular metrics}. Generated function samples are presented in Figure \ref{fig:Volcano/gano_samples}. We present reconstructions from the data-set in Figure \ref{fig:Volcano/test_reconstructions} and new generated function samples in Figure \ref{fig:Volcano/vano_samples}.

\section{Comparison Metrics}
\label{sec: metrics}

In this section we provide a description of different metrics used for evaluating the quality of our results. 

\subsection{Hilbert-Schmidt norm} \label{sec: HS norm}
The Hilbert-Schmidt norm of an operator $T:\curlyH \to \curlyH$ on a Hilbert space $\curlyH$ with orthonormal basis $e_i$ is given by
\begin{equation}
\|T\|_{HS}^2 = \sum_{i} \langle Te_i, e_i \rangle.
\end{equation}
If the operator $T$ is self-adjoint with eigenvalues $\lambda_i$, this can also be written as
\begin{equation}
\|T\|_{HS}^2 = \sum_i \lambda_i^2.
\end{equation}
Note that when $\curlyH$ is a finite dimensional Hilbert space, this is equivalent to the standard 2 (Frobenius) norm for operators (matrices).

Since covariance operators for Gaussian measures always have finite Hilbert-Schmidt norm, we measure the distance of the two mean-zero Gaussian measures in the Gaussian random field example via the Hilbert-Schmidt norm of their difference.  We approximate this via the approximations of the covariance operators in the discretization space $\R^{128 \times 128}$,
\begin{equation}
C = \sum_{i=1}^{n_{eig}} \lambda_i \tilde \varphi_i \tilde \varphi_i^\top, \quad \hat C = \sum_{i=1}^n \tilde \tau_i \tilde \tau_i^\top,
\end{equation}
where $\tilde \phi_i, \tilde \tau_i \in \R^{128}$ are the evaluations of the functions $\phi_i$ and $\tau_i$ along the measurement points used in the experiment.  The normalized Hilbert-Schmidt norm of the difference of the true covariance operators is then approximated as the Frobenius norm of the difference of their approximations divided by the Frobeinus norm of the true covariance $C$.

\subsection{Generalized Maximum Mean Discrepancy}
\label{sec: generalized mmd}

For measuring the distance between ground truth and learned distributions, we choose to use a version of the Maximum Mean Discrepancy (MMD) distance.  Given a probability distribution on a set $\curlyX$ and a characteristic kernel $k: \curlyX \times \curlyX \to \R$ \cite{sriperumbudur2011universality}, the \emph{kernel mean embedding} \cite{muandet2017kernel} is a map from probability measures $\mu$ on $\curlyX$ into the Reproducing Kernel Hilbert Space (RKHS) associated with $k$, $\curlyH_k$ given by
\begin{equation}
\hat \mu_k := \int_\curlyX k(\cdot, x) \d \mu(x).
\end{equation}
Note that if $\mu$ is an empirical distribution, that is, a sum of delta measures
\begin{equation*}
\mu = \frac{1}{N}\sum_{i=1}^N \delta_{x_i},
\end{equation*}
then the kernel mean embedding is given by
\begin{equation}
\hat \mu_k = \frac{1}{N} \sum_{i=1}^N k(\cdot, x_i).
\end{equation}
Given two probability measures, $\mu$ and $\nu$ on a set $\curlyX$, we can define the MMD distance between them as the distance between their kernel mean embeddings in the RKHS $\curlyH_k$,
\begin{equation}
\mathrm{MMD}_k(\mu,\nu) = \|\hat\mu_k - \hat \nu_k\|_{\curlyH_k}^2.
\end{equation}
When both $\mu$ and $\nu$ are empirical distributions on points $\{x_i\}_{i=1}^N$ and $\{y_j\}_{j=1}^M$, respectively, the MMD can be evaluated as
\begin{equation}
\|\hat\mu_k - \hat \nu_k\|_{\curlyH_k}^2 = \frac{1}{N^2} \sum_{i,k = 1}^N k(x_i, x_k) + \frac{1}{M^2} \sum_{j,\ell = 1}^M k(y_j, y_\ell) - \frac{2}{NM} \sum_{i=1}^N\sum_{j=1}^M k(x_i,y_j).
\end{equation}

While this is convenient for giving a notion of distance between empirical distributions corresponding to samples from a data-set and a generative model, it can be sensitive to the form of the kernel.  For example, if a norm on $\curlyX$ is used in a Gaussian kernel with a length-scale $\sigma$,
\begin{equation}
k_\sigma(x,y) = \exp\left(\frac{1}{2\sigma^2} \|x - y\|_\curlyX^2\right),
\end{equation}
for large enough $\sigma$ the kernel will see all data points as being roughly the same and the MMD for any two fixed empirical distributions will become arbitrarily small.

To mitigate this problem, the generalized MMD distance was proposed \cite{fukumizu2009kernel}, which instead of using a single kernel uses a family of kernels $\curlyF$ and defines a (pseudo-)metric between probability measures as
\begin{equation}
\mathrm{GMMD}_\curlyF(\mu,\nu) := \underset{k \in \curlyF}{\sup} \;\;\mathrm{MMD}_k(\mu, \nu).
\end{equation}
As long as one of the kernels in $\curlyF$ is characteristic, this defines a valid distance \cite{fukumizu2009kernel}.

\begin{figure}
\begin{center}
\centerline{\includegraphics[width=0.35\textwidth]{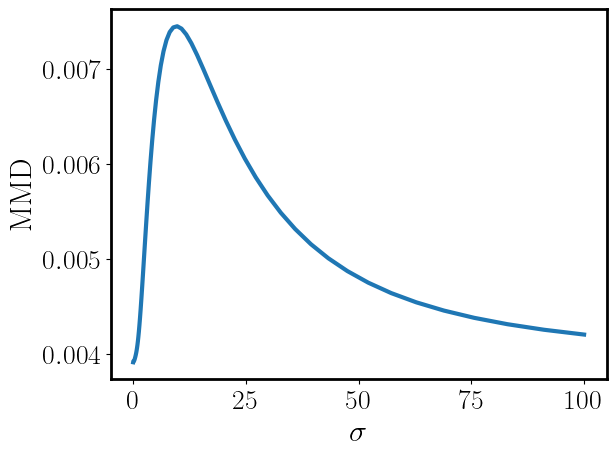}}
\caption{An example of MMDs computed over a range of lengthscales $\sigma$ between the ground truth 2D Gaussian density data-set and the data-set generated by VANO.}
\label{fig:GaussianBump/generalized_mmd}
\end{center}
\end{figure}

In our experiments, we use the GMMD as a measure of distance of distributions with the family of kernels 
\begin{equation}
\curlyF = \left\{ k_\sigma\;|\;k_\sigma(x,y) = \exp\left(\frac{1}{2\sigma^2} \|x - y\|_\curlyX^2\right),\;\sigma_- \leq \sigma \leq \sigma_+\right\}.
\end{equation}
Empirically, we find that the $\sigma$ giving the largest $MMD$ lies within the interval $\sigma_- = .1$ and $\sigma_+= 20$ for all experiments, and therefore use a mesh of $\sigma$ in this interval to approximate this $GMMD$.  In Figure \ref{fig:GaussianBump/generalized_mmd} we plot an example of the MMD for varying $\sigma$ between 512 function samples from the 2D Gaussian densities data-set and those generated from the VANO model.  

\subsection{Circular variance and Skewness}
\label{sec: circular metrics}

The circular variance and skewness are moments of circular random variables, see \cite{rahman2022generative}, used to evaluate the quality of generated angular valued functions. For $N$ random variables given by angles, $\{\theta_j\}_{j=1}^N$, let $z_p = \sum_i^{N} e^{i p \theta_j}$ with $i = \sqrt{-1}$.  Define $\varphi_p = \mathrm{arg}(z_p)$ where $\arg$ is the complex argument function (returns the angle of a complex number to the real axis) and let $R_p = |z_p|/N$.
The circular variance is then defined by $\sigma = 1 - R_1$ and the skewness by $s = \frac{R_2 \sin(\varphi_2 - 2 \varphi_1)}{(1- R_1)^{3/2}}$.

\section{Trainable Parameters and Computational Cost}
\label{sec:par time}

We present the training time in seconds for each experiment and model in the manuscript in Table \ref{tab:computational_cost} as well as the total number of trainable parameters in Table \ref{tab:number_of_parameters}.

\begin{table}
\caption{Computational cost for training  all the models considered in this manuscript: We present the wall clock time \emph{in seconds} that is needed to train each model on a single NVIDIA RTX A6000 GPU.}
\label{tab:computational_cost}
\begin{center}
\begin{small}
\begin{sc}
\begin{tabular}{lcccr}
\toprule
Benchmark & VAE & VANO (linear) & VANO (nonlinear) & GANO \\
\midrule
GRF & - & 53 & - & - \\
\midrule
2D Gaussian Densities & - & 67 &  198 & - \\
\midrule
 Cahn-Hilliard ($64 \times 64$) & 43 & - & 1,020 & - \\
\midrule
Cahn-Hilliard ($128 \times 128$) & 55  & -  & - & - \\
\midrule
 Cahn-Hilliard ($256 \times 256$)& 166  & -  & -  & -\\
 \midrule
InSAR Interferogram & - & - & 11,820 & 42,060 \\
\bottomrule
\end{tabular}
\end{sc}
\end{small}
\end{center}
\end{table}

\begin{table}
\caption{Total number of trainable parameters for all the models considered in this manuscript.}
\label{tab:number_of_parameters}
\begin{center}
\begin{small}
\begin{sc}
\begin{tabular}{lcccr}
\toprule
Benchmark & VAE & VANO (linear) & VANO (nonlinear) & GANO \\
\midrule
 GRF  & - & 107,712 & - & -  \\
 \midrule
 2D Gaussian Densities & - & 85,368 & 89,305 & -  \\
\midrule
 Cahn-Hilliard ($64 \times 64$) & 187,000 & - & 341,000 & -  \\
\midrule
  Cahn-Hilliard ($128 \times 128$) & 485,000 & -  & - & -  \\
\midrule
 Cahn-Hilliard ($256 \times 256$)& 1,667,000 & -  & - & - \\
 \midrule
 InSAR Interferogram & - & - & 11,130,420 & 48,827,763  \\
\bottomrule
\end{tabular}
\end{sc}
\end{small}
\end{center}
\end{table}

\end{document}